\documentclass[acmtog]{acmart}
\setcitestyle{square} 
\setcopyright{acmlicensed}
\acmJournal{TOG}
\acmYear{2018}\acmVolume{37}\acmNumber{4}\acmArticle{95}\acmMonth{8} \acmDOI{10.1145/3197517.3201304}

\usepackage{amsmath}
\usepackage{lineno}
\usepackage{graphicx}
\usepackage{subcaption}
\usepackage{calc}
\usepackage{booktabs}
\usepackage{tabularx}
\usepackage[english]{babel}
\usepackage{booktabs}
\usepackage{overpic}
\usepackage{breqn}
\usepackage{multirow}
\usepackage{algorithm}
\usepackage[noend]{algpseudocode}
\usepackage{dblfloatfix}
\usepackage{color}
\usepackage{xcolor}
\usepackage[
singlelinecheck=false
]{caption}

\definecolor{green}{rgb}{0.125,0.65,0.125}
\definecolor{reddish}{rgb}{0.7,0.,0.}
\definecolor{orange}{rgb}{1.0,0.3,0.1}
\definecolor{orange2}{rgb}{0.6,0.1,0.1}
\definecolor{water}{rgb}{0.25,0.35,0.75}
\definecolor{purple2}{rgb}{1.0,0.0,1.0}

\definecolor{blue1}{rgb}{0.1,0.1,0.3}
\definecolor{green1}{rgb}{0.5,1.0,0.82}
\usepackage{soul}

\newcommand{\revision}[1]{{\color{black}#1}}

\newcommand{\newRachel}[1]{{{#1}}}

\newcommand\Tstrut{\rule{0pt}{2.6ex}}
\newcommand\Tstruta{\rule{0pt}{4ex}}
\newcommand{\resizeEq}[2]{\begin{equation}
\resizebox{0.92\hsize}{!}{
$\begin{array}{@{}ll}
\begin{split}
#1
\end{split}
\end{array}$}  \label{#2}
\end{equation}}

\newcommand{\myrefeq}[1]{Eq.~\eqref{#1}}
\newcommand{\myreffig}[1]{Fig.~\ref{#1}}

\newcommand{\myrefsec}[1]{Sec.~\ref{#1}}
\newcommand{\myrefapp}[1]{Appendix~\ref{#1}}
\newcommand{\myrefalg}[1]{Alg.~\ref{#1}}

\makeatletter
\renewcommand{\ALG@name}{Alg.}
\makeatother

\newcommand{\vel}{\mathbf{v}}
\newcommand{\vort}{\mathbf{w}}
\newcommand{\dens}{\rho}
\renewcommand{\vec}{\mathbf}
\newcommand{\myavg}{\mathbb{E}}
\begin{abstract}
We propose a temporally coherent generative model addressing 
the super-resolution problem for fluid flows. 
Our work represents a first approach
to synthesize four-dimensional physics fields with neural networks.
Based on a conditional generative adversarial network
that is designed for the inference of three-dimensional volumetric data, 
our model generates consistent and detailed results 
by using a novel temporal discriminator, in addition to the commonly used spatial one.
Our experiments show that the generator is able to
infer more realistic high-resolution details by using
additional physical quantities, such as low-resolution velocities or vorticities.
Besides improvements in the training process and in the generated outputs, 
these inputs offer means for artistic control as well.
We additionally employ a physics-aware data augmentation step,
which is crucial to avoid overfitting and to reduce memory requirements.
In this way, our network learns to 
generate advected quantities with highly detailed, realistic, and temporally coherent features.
Our method works instantaneously, using only a single time-step of low-resolution fluid data.
We demonstrate the abilities of our method using a variety of complex inputs and applications in
two and three dimensions.
\end{abstract}

\begin{document}
\title{tempoGAN: A Temporally Coherent, Volumetric GAN for Super-resolution Fluid Flow}
\author{You Xie*}
\affiliation{
	\institution{Technical University of Munich}}
\email{you.xie@tum.de}
\author{Erik Franz*}
\affiliation{
	\institution{Technical University of Munich}}
\email{franzer@in.tum.de}
\author{Mengyu Chu*}
\affiliation{
	\institution{Technical University of Munich}}
\email{mengyu.chu@tum.de}
\author{Nils Thuerey}
\affiliation{
	\institution{Technical University of Munich}}
\email{nils.thuerey@tum.de}
\renewcommand\shortauthors{Xie, Y., Franz, E., Chu, M., Thuerey, N.}

\ccsdesc[500]{Computing methodologies~Neural networks}
\ccsdesc[500]{Computing methodologies~Physical simulation}

\keywords{physics-based deep learning, generative models, computer animation, fluid simulation}

\thanks{
(*) Similar amount of contributions.
 }

\begin{teaserfigure}
	\centering
	\includegraphics[width=\linewidth]{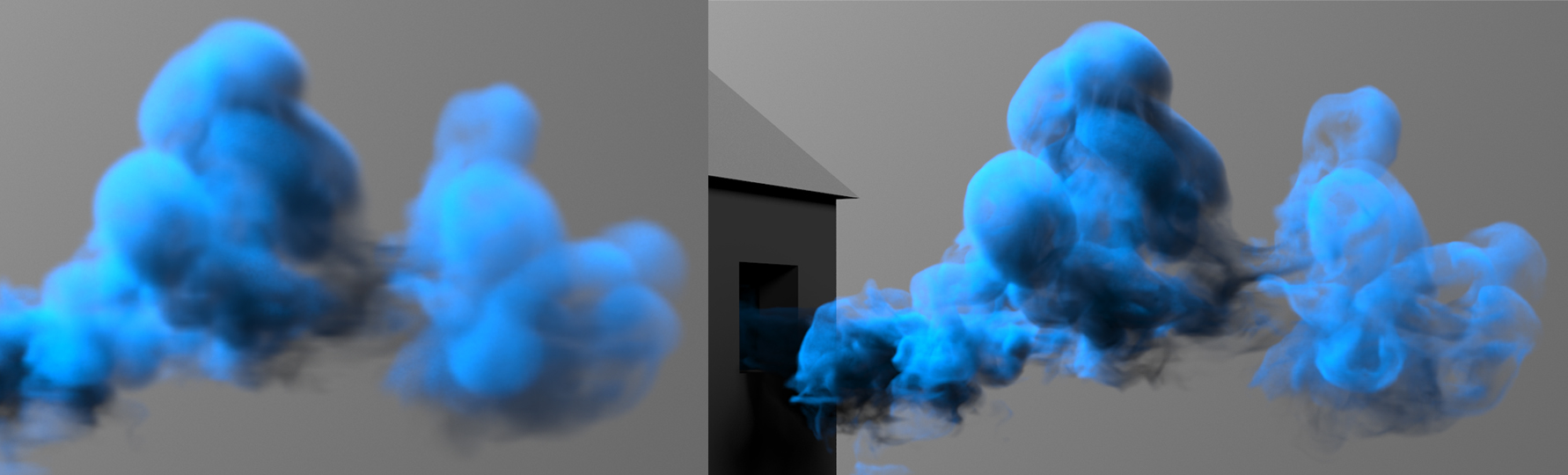}
	\caption{Our convolutional neural network learns to generate highly detailed,
	and temporally coherent features based on a low-resolution field containing a single time-step
	of density and velocity data. We introduce a novel discriminator that ensures
	the synthesized details change smoothly over time.}
	\label{fig:teaser}
\end{teaserfigure}

\maketitle

\section{Introduction} \label{sec:intro}

Generative models were highly successful in the last years
to represent and synthesize complex natural images 
\cite{goodfellow2014generative}.
These works demonstrated that deep convolutional neural networks (CNNs) are able to
capture the distribution of, e.g., photos of human faces, and generate novel,
previously unseen versions that are virtually indistinguishable from the original
inputs.  Likewise, similar algorithms were shown to be extremely successful at
generating natural high-resolution images from a coarse input \cite{karras2017growgan}. 
However, in their original
form, these generative models do not take into account the temporal
evolution of the data, which is crucial for realistic physical systems. In the
following, we will extend these methods to generate high-resolution volumetric
data sets of passively advected flow quantities, and ensuring temporal coherence 
is one of the core aspects that we will focus on below.
We will demonstrate that it is especially important to make the training process
aware of the underlying transport phenomena, such that the network can learn to
generate stable and highly detailed solutions.

Capturing the intricate details of turbulent flows has been a long-standing
challenge for numerical simulations. Resolving such details with discretized
models induces enormous com\-pu\-tatio\-nal costs and quickly becomes
infeasible for flows on human space and time scales. 
While algorithms to increase the apparent resolution
of simulations can alleviate this problem \cite{Kim:2008:wlt}, 
they are typically based on procedural models that are only loosely inspired by
the underlying physics.
In contrast to all previous methods, our algorithm represents a
physically-based interpolation, that does not require any form
of additional temporal data or quantities tracked over time. The
super-resolution process is instantaneous, based on volumetric
data from a single frame of a fluid simulation.
We found that inference of 
high-resolution data in a fluid flow setting benefits from the availability
of information about the flow. In our case, this takes the shape of
additional physical variables such as velocity and vorticity
as inputs, which in turn yield means for artistic control.
A particular challenge in the field of super-resolution flow is how to 
evaluate the quality of the generated output. As we are typically
targeting turbulent motions, a single coarse approximation can be associated
with a large variety of significantly different high-resolution versions.
As long as the output matches the correlated spatial 
and temporal distributions of the reference data, it represents a correct solution. To encode this requirement in the training
process of a neural network, we employ so-called generative adversarial
networks (GANs). These methods
train a {\em generator}, as well as a second network, the {\em discriminator} 
that learns to judge how closely the generated output matches the ground truth data.
In this way, we train a specialized, data-driven loss function
alongside the generative network, while making sure it
is differentiable and compatible with the training process.
We not only employ this adversarial approach for the
smoke density outputs, but we also train a specialized and novel adversarial
loss function that learns to judge the temporal coherence of the
outputs. 

We additionally present best practices to set up a training pipeline for
physics-based GANs. E.g., we found it particularly useful to have physics-aware
data augmentation functionality in place. The large amounts of space-time data
that arise in the context of many physics problems quickly bring typical
hardware environments to their limits. 
As such, we found data augmentation crucial to avoid overfitting.
We also explored a variety of different variants
for setting up the networks as well as training them, 
and we will evaluate them in terms of their capabilities to learn
high-resolution physics functions below.

\begin{figure}[tb]
	\centering 
	\includegraphics[width=\linewidth]{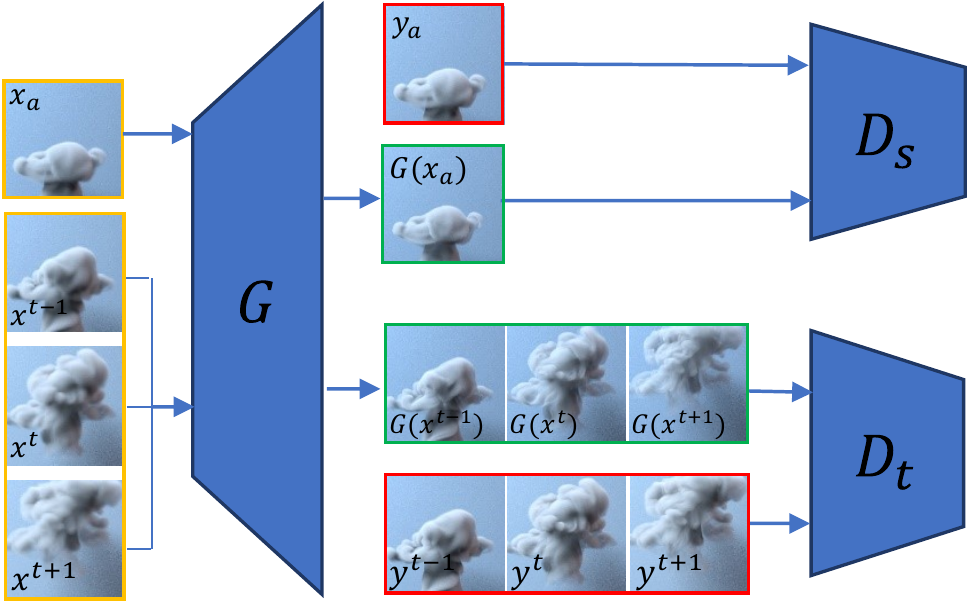} \\
	\caption{ \revision{This figure gives a high level overview of our approach: a generator on the 
	left, is guided during training by two discriminator networks (right), one of which
	focuses on space ($D_s$), while the other one focuses on temporal aspects ($D_t$).
	At runtime, both are discarded, and only the generator network is evaluated.} }
	\label{fig:tempoGANcoarse}
\end{figure}

To summarize, the main contributions of our work are:
\begin{itemize}
{\vspace{-2pt}
\item a novel temporal discriminator, to generate consistent and high\-ly detailed results over time,
\item artistic control of the outputs, in the form of additional loss terms and
	an intentional entangling of the physical quantities used as inputs,
\item a physics aware data augmentation method,
\item and a thorough evaluation of adversarial training processes for physics functions.}
\end{itemize}
To the best of our knowledge, our approach is the first generative adversarial network
for four-dimensional functions,
and we will demonstrate that it
successfully learns to infer solutions for flow transport processes from approximate solutions.
\revision{ A high level preview of the architecture we propose can
be found in \myreffig{fig:tempoGANcoarse}.}
\section{Related Work}
In the area of computer vision, deep learning techniques have achiev\-ed significant breakthroughs
in numerous fields such as classification \cite{krizhevsky2012imagenet}, object detection \cite{girshick2014rich}, 
style transfer \cite{luan2017deep}, novel view synthesis \cite{flynn2016deepstereo}, and 
additionally, in the area of content creation. For more in-depth reviews of neural networks
and deep learning techniques, we refer the readers to corresponding books \cite{bishop2006book,Goodfellow2016}. 

One of the popular methods to generate content are so called {\em generative adversarial networks} (GANs), 
introduced by Goodfellow et al. \cite{goodfellow2014generative}. They were shown 
to be particularly powerful at re-creating the distributions of complex data sets such as images
of human faces.
Depending on the kind of input data they take, GANs can be separated into unconditional and conditional ones. 
The formers generate realistic data from samples of a synthetic data distribution
like Gaussian noise. The DC-GAN \cite{RadfordMC15} is a good example of an unconditional
GAN. It was designed for generic natural images, 
while the cycle-consistent GAN by Zhu et al.  \shortcite{zhu2017cycle} was developed
to translate between different classes of images.
The conditional GANs were introduced by Mirza and Osindero \shortcite{mirza2014conditional}, and
provide the network with an input that is in some way related to the target function in order
to control the generated output.
Therefore, conditional variants are popular for transformation tasks,
such as image translations problems \cite{isola2016image} and 
super resolution problems \cite{ledig2016photo}. 

In the field of super-resolution techniques, researchers have explored different network architectures.
E.g., convolutional net\-works \cite{dong2016image} were shown to be more effective than fully connected architectures.
These networks can be trained with smaller tiles and
 later on be applied to images of arbitrary sizes \cite{isola2016image}.
Batch normalization \cite{lim2017enhanced} significantly improves results by removing value 
shifting in hidden layers, 
and networks employing so-called {\em residual blocks}
\cite{kim2016accurate, lim2017enhanced} enable the training of deep networks 
without strongly vanishing gradients.

In term of loss functions, pixel-wise loss between the network output and the ground truth data 
used to be common, such as the $L_1$ and $L_2$ loss \cite{dong2016image}.
Nowadays, using the adversarial structure of GANs, or using pre-trained
networks, such as the VGG net \cite{simonyan2014} often leads to higher perceptual qualities
\cite{mathieu2015deep, johnson2016perceptual}.

Our method likewise uses residual blocks in conjunction with a conditional GAN architecture to
infer three-dimensional flow quantities. Here, we try to use standard architectures. 
While our generator is similar
to approaches for image super-resolution \cite{ledig2016photo}, 
we show that loss terms and discriminators are crucial for high-quality outputs.
We also employ a fairly traditional GAN training, instead of recently proposed alternatives
\cite{arjovsky2017wasserstein,berthelot2017began}, which could potentially lead to additional gains in quality.
Besides the super-resolution task, our work differs from many works in the GAN area
with its focus on temporal coherence, as we will demonstrate in more detail later on.

While most works have focused on single images, several papers have addressed
temporal changes of data sets. One way to solve this problem is by 
directly incorporating the time axis, i.e., by using sequences of data as input and output.
E.g., Saito et al. propose a temporal generator in their work \cite{saito2017temporal},
while Yu et al. \cite{yu2017seqgan} proposed a sequence generator that learns a stochastic policy.
In these works, results need to be generated
sequentially, while our algorithm processes individual frames independently, and in arbitrary order, if necessary.
In addition, such approaches would explode in terms of weights and computational resources for
typical four-dimensional fluid data sets.

An alternative here is to generate single frame data with additional
loss terms to keep the results coherent over time.
Bhattacharjee etc. \shortcite{bhattacharjee2017temporal} achieved improved coherence in their results for video frame prediction, 
by adding specially designed distance measures as a discontinuity penalty between nearby frames.
For video style transfer, a $L_2$ loss on warped nearby frames helped to 
alleviate temporal discontinuities, as shown by Ruder et al. \shortcite{Ruder2016Artistic}. 
In addition to a $L_2$ loss on nearby frames, Chen et al. \shortcite{Chen2017ICCV} used neural networks to learn 
frame warping and frame combination in VGG feature space.
Similarly, Liu etc. \cite{liu2017VideoSR} used neural networks to learn
spatial alignment for low-resolution inputs, and adaptive aggregation for high-resolution outputs,
which also improved the temporal coherence.
Due to the three-dimensional data sets we are facing, we also adopt the single
frame view. However, in contrast to all previous works, we propose the use
of a temporal discriminator. We will show that relying on data-driven, learned
loss functions in the form of a discriminator helps to improve results over manually
designed losses. Once our networks are trained, this discriminator can be discarded.
Thus, unlike, e.g., aggregation methods, our approach does not influence runtime performance.
While previous work shows that warping layers are useful in motion field 
learning~\cite{de2017deep, Chen2017ICCV}, our work targets the opposite direction:
by providing our networks with velocities, warping layers 
can likewise improve the training of temporally coherent content generation.

More recently, deep learning algorithms have begun to influence 
computer graphics algorithms.
E.g., they were successfully used for efficient and noise-free renderings \cite{bako2017kernel,chaitanya2017interactive}, 
the illumination of volumes \cite{kallweit2017deep}, 
for modeling porous media \cite{mosser2017porous}, and for character control \cite{peng2017deeploco}.
First works also exist that target numerical simulations. E.g., a conditional GAN
was used to compute solutions for smaller, two-dimensional advection-diffusion
problems \cite{farimani2017, long2017pde}.
Others have demonstrated the inference of SPH forces with regression forests \cite{ladicky2015data},
proposed CNNs for fast pressure projections \cite{tompson2016accelerating},
learned space-time deformations for interactive liquids \cite{rtliquids2017} , and 
modeled splash statistics with NNs \cite{um2017mlflip}.
Closer to our line of work, Chu et al. \shortcite{chu2017cnnpatch} proposed
a method to look up pre-computed patches using CNN-based descriptors. Despite a similar goal, 
their methods still require additional Lagrangian tracking information, while our method
does not require any modifications of a basic solver. In addition, our method does not use
any stored data at runtime apart from the trained generator model.

As our method focuses on the conditional inference of
high-re\-so\-lu\-tion flow data sets, i.e. solutions of the {\em Navier-Stokes} (NS) equations, we also give a brief overview of the
related work here, with a particular focus on single-phase flows.
\revision{After the introduction of the stable fluids algorithm \cite{Stam1999},} 
\revision{a variety of extensions and variants have been developed over the years.
E.g., more accurate Eulerian ad\-vec\-tion sche\-mes \cite{Kim05FlowFixer,Selle:2008:USM} 
are often employed, an alternative to which are Lagrangian versions \cite{rasmussen2003smoke,magnus2011capturing}.
While grids are more commonly used, particles can achieve non-dissipative results for which
a Eulerian grid would require a significant amount of refinement.
Furthermore, procedural turbulence methods to increase apparent resolutions are popular 
extensions \cite{Kim:2008:wlt,narain:2008:procTurb,schechter2008evolving}.
In contrast to our work, the different advection schemes and procedural
turbulence methods require a calculation of the
high-resolution transport of the density field over the full simulation sequence. 
Additionally, Eulerian and Lagrangian representations can be advected in a parallelized 
fashion on a per frame basis, in line with the application of convolutions for NNs.
Our method infers an instantaneous solution to the underlying advection problem
based only on a single snapshot of data, without having to compute a series of previous 
time steps.}

Our work also shares similarities in terms
of goals with other phy\-sics-based up-sampling algorithms \cite{kavan2011physics},
and due to this goal, is related to fluid control methods \cite{McNamaraAdjointMethod,Pan:2013}.
These methods would work very well in conjunction with our approach, in order to 
generate a coarse input with the right shape and timing.

\section{Adversarial Loss Functions}
\label{sec:loss}
Based on a set of low-resolution inputs, with corresponding
high-resolution references, our goal is to train a CNN that produces
a temporally coherent, high-resolution solution with adversarial train\-ing.
We will first very briefly summarize the basics of adversarial training,
and then explain our extensions for temporal coherence and for \revision{results control.} 
\subsection{Generative Adversarial Networks}
\label{sec:ganbasic}
GANs consist of two models, which are trained in conjunction:
the generator $G$ and the discriminator $D$. Both will be realized as convolutional neural networks in our case. 
For regular ones, i.e., not conditional GANs,
the goal is to train a generator $G(x)$ that maps a simple data distribution, typically noise, $x$ to a complex desired output $y$, e.g., natural images.
Instead of using a manually specified loss term to train the generator, another NN, the discriminator, is used as complex, learned loss function \cite{goodfellow2014generative}.
This discriminator takes the form of a simple binary classifier, which is trained in a supervised manner 
to reject {\em generated} data, i.e., it should return $D(G(x)) = 0$, and accept the {\em real} data with $D(y) =1$.
For training, the loss for the discriminator is thus given by a sigmoid cross entropy
for the two classes ``generated'' and ``real'':
\revision{
\resizeEq{
\begin{split}
\mathcal{L}_D(D,G)= & \mathbb{E}_{y\sim p_{\text{y}}(y)}[-\log D(y)] +\mathbb{E}_{x\sim p_{\text{x}}(x)}[-\log (1-D(G(x)))] \\
= & \myavg_m [-\log D(y_m)] + \myavg_n [-\log (1-D(G(x_n)))]\ ,
\end{split} 
}{eq:disloss}
}
where $n$ is the number of drawn inputs $x$, while $m$ denotes the number of real data samples $y$.
\revision{Here we use the notation $y\sim p_{\text{y}}(y)$ for samples $y$ being drawn from a corresponding 
probability data distribution $p_{\text{y}}$, which will later on be represented by our numerical simulation framework.}
\revision{The continuous distribution $\mathcal{L}_D(D,G)$ yields the average of discrete samples $y_n$ and $x_m$ in the second line of \myrefeq{eq:disloss}.
We will omit the $y\sim p_{\text{y}}(y)$ and $x\sim p_{\text{x}}(x)$ subscripts of the sigmoid cross entropy,}
and $n$ and $m$ subscripts of $D(y_m)$ and $G(x_n)$, for clarity below.

In contrast to the discriminator, the generator is trained to ``fool'' the discriminator into accepting its samples and thus to generate output that is close to the real data from $y$.
In practice, this means that the generator is trained to drive the discriminator result for its outputs to one.
Instead of directly using the negative discriminator loss, GANs typically use
\revision{
\resizeEq{
\begin{split}
\mathcal{L}_G(D,G)= \mathbb{E}_{x\sim p_{\text{x}}(x)}[-\log (D(G(x)))] = \myavg_n [-\log (D(G(x)))]
\end{split}}{}
}
as the loss function for the generator, in order to reduce diminishing gradient problems 
\cite{goodfellow2016nips}.
As $D$ is realized as a NN, it is guaranteed to be sufficiently differentiable as a loss function for $G$.
In practice, both discriminator and generator are trained in turns and will optimally reach an
equilibrium state.

As we target a super-resolution problem, our goal is not to generate an arbitrary high-resolution output, but
one that corresponds to a low-resolution input, and hence we employ a {\em conditional} GAN.
In terms of the dual optimization problem described above, this means that the input $x$ now represents
the low-resolution data set, and the discriminator is provided with $x$ in order to establish and 
ensure the correct relationship
between input and output, i.e., we now have $D(x,y)$ and $D(x,G(x))$ \cite{mirza2014conditional}.
Furthermore, previous work \cite{zhao2015loss}
has shown that an
additional $L_1$ loss term with a small weight can be added to the generator to ensure that its output stays close 
to the ground truth $y$. This yields $\lambda _{L_1}  \myavg_n \left \| G(x)-{y}\right \|_{1}$, where $\lambda _{L_1}$
controls the strength of this term, and we use $\myavg$ for consistency to denote the
expected value, in this discrete case being equivalent to an average.

\subsection{ Loss in Feature Spaces}
\label{sec:featureloss}

\revision{In order to further control the coupled, non-linear optimization process, the features
of the underlying CNNs can be constrained. This is an important issue, as controlling
the training process of GANs is known as a difficult problem.
Here, we extend previous work on feature space losses, which
were shown to improve realism in natural images~\cite{dosovitskiy2016generating}, and
were also shown to help with mode collapse problems \cite{salimans2016improved}.} 
To achieve this goal, an $L_2$ loss over parts or the whole feature space of a neural network is introduced for the generator.
I.e., the intermediate results of the generator network are constrained w.r.t. a set of intermediate reference data.
While previous work typically makes use of manually selected layers
of pre-trained networks, such as the VGG net, we propose to use features of the discriminator as constraints instead.

Thus, we incorporate a novel loss term of the form
\begin{equation}
\mathcal{L}_{f}=\myavg _{n,j} \lambda _{f}^{j}\left \| F^{j}(G(x))-F^{j}(y) \right \|_{2}^{2} \ ,
\label{eq:featureloss}
\end{equation}
where $j$ is a layer in our discriminator network, and $F^{j}$ denotes the activations of the corresponding layer.
The factor $\lambda _{f}^{j}$ is a weighting term, which can be adjusted on a per layer basis, as we will discuss in \myrefsec{sec:artcontrol}. 
It is particularly important in this case that we can employ the discriminator here, as no suitable, pre-trained networks
are available for three-dimensional flow problems.

\begin{figure}[t]
     \centering 
	 \begin{overpic}[width=0.24\linewidth]{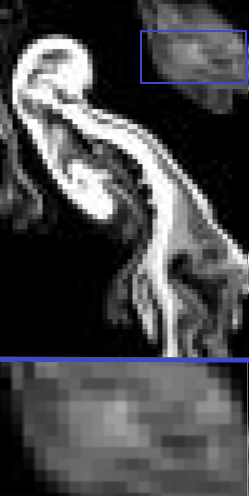}
	 \put( 5,5){\small \color{white}{$a)$}}
	 \end{overpic}
	 \begin{overpic}[width=0.24\linewidth]{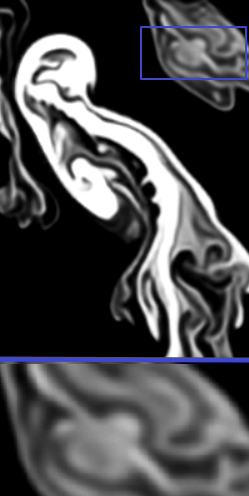}
	 \put( 5,5){\small \color{white}{$b)$}}
 	\end{overpic}
	 \begin{overpic}[width=0.24\linewidth]{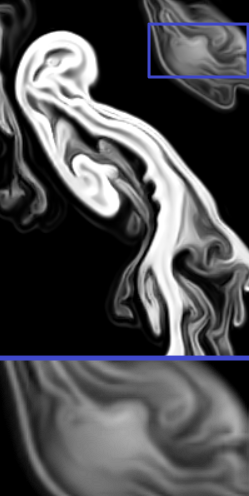}
	 \put( 5,5){\small \color{white}{$c)$}}
	 \end{overpic}
	 \begin{overpic}[width=0.24\linewidth]{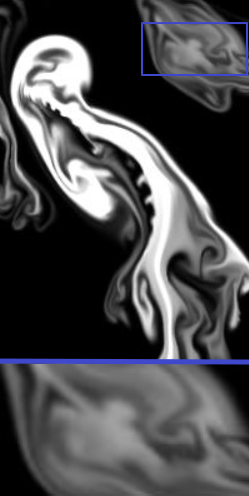}
	\put( 5,5){\small \color{white}{$d)$}}
   	\end{overpic}
     \caption{From left to right: a) a sample, low-resolution input, b) a CNN output with naive $L_2$ loss (no GAN training), c)
     our tempoGAN output, and d) the high-resolution reference. The $L_2$ version learns a smooth result
     without small scale details, while our output in (c) surpasses the detail of the reference in certain regions.
     } \label{fig:2dlayerlossexample}
\end{figure}

\revision{Interestingly, these weights yields different and realistic results both for positive as well as negative choices for the weights.
For $\lambda _{f} > 0$
these loss terms effectively encourage a minimization of the mean feature space distances of real and
generated data sets, such that generated features resemble features of the reference. 
Surprisingly, we found that training runs with $\lambda _{f} < 0$
also yield excellent, and often slightly better results.
As we are targeting conditional GANs, our networks
are highly constrained by the inputs. 
Our explanation for this behavior is that a negative feature loss in this setting encourages the optimization
to generate results that differ in terms of the features, but are still similar, ideally indistinguishable, in 
terms of their final output. This is possible as we are not targeting a single ground-truth result, but 
rather, we give the generator the freedom to generate any result that best fits the collection of inputs
it receives. From our experience, this loss term drives the generator towards realistic detail,
an example of which can be seen in \myreffig{fig:2dlayerlossexample}. 
Note 
that due to the non-linear nature of the optimization, linearly changing $\lambda _{f}$ 
yields to models with significant differences in the generated small scale features.}

\subsection{ Temporal Coherence}
\label{sec:tempo}
While the GAN process described so far is highly successful at generating highly 
detailed and realistic outputs for static frames, these details are particularly challenging
in terms of their temporal coherence.
Since both the generator and the discriminator work on every frame independently, subtle changes of the input $x$ can lead to outputs $G(x)$ with distinctly different 
 details for higher spatial frequencies.

When the ground truth data $y$ comes from a transport process, such as frame motion or flow motion, it
 typically exhibits a very high degree of temporal coherence, and a velocity field $v_y$ exists for which 
 $y^{t} = \mathcal{A}( y^{t-1}, v_y^{t-1} )$. Here, we denote the advection operator
 (also called warp or transport in other works) with
$\mathcal{A}$, and we assume without loss of generality
that the time step between frame $t$ and $t-1$ is equal to one. Discrete time steps
will be denoted by superscripts, i.e., for a function $y$ of space and time $y^t = y(\vec{x},t)$ denotes
a full spatial sample at time $t$.
Similarly, in order to solve the temporal coherence problem, the relationship $G(x^{t}) = \mathcal{A}( G(x^{t-1}), v_{G(x)}^{t-1})$ should hold, 
which assumes that we can compute a motion $v_{G(x)}$ based on the generator input $x$. 
While directly computing such a motion can be difficult and unnecessary for general GAN problems, 
we can make use of the ground truth data for $y$ in our conditional setting. I.e., in the following,
we will use a velocity reference $v_y$ corresponding to the target $y$, and perform a 
spatial down-sampling to compute the velocity $v_x$ for input $x$.

Equipped with $v_x$, one possibility to improve temporal coherence would be
to add an $L_2$ loss term of the form:
\begin{equation}\label{eq:l2VelSingleSide}
\mathcal{L}_{2,t} = \| G(x^{t}) - \mathcal{A}( G(x^{t-1}), v_{x}^{t-1})\|_{2}^{2}
\end{equation}
We found that extending the forward-advection difference with backward-advection improves the results further, i.e., the 
following $L_2$ loss is clearly preferable over \myrefeq{eq:l2VelSingleSide}:
\resizeEq{
\begin{split}
\mathcal{L}_{2,t} = \| G(x^{t}) - \mathcal{A}( G(x^{t-1}), v_{x}^{t-1})\|_{2}^{2} + \| G(x^{t}) - \mathcal{A}( G(x^{t+1}), -v_{x}^{t+1})\|_{2}^{2} \
\end{split}
}{eq:l2VelDoubleSide}
, where we align the next frame at $t+1$ by advecting with $-v_{x}^{t+1}$.

\begin{figure}[b]
	\centering
	\begin{overpic}[width=\linewidth]{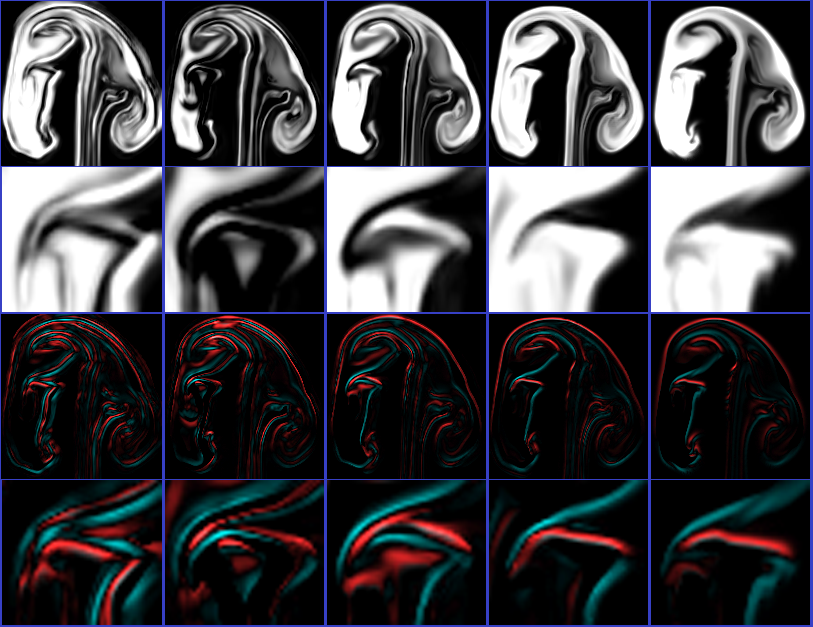}
		\put( 1.2,1.1){\small \color{white}{None}}
		\put( 21,2){\small \color{white}{$\mathcal{L}_{2,t}$}}
		\put( 41,2){\small \color{white}{$\mathcal{L}_{ \footnotesize{D_t}' }$}}
		\put( 61,2){\small \color{white}{$\mathcal{L}_{D_t}$}}
		\put( 81,2){\small \color{white}{$y$}}
	\end{overpic}
	\caption{ A comparison of different approaches for temporal coherence.
		The top two rows show the inferred densities, while the bottom two rows contain the 
		time derivative of the frame content computed with a finite difference between frame $t$ and $t+1$.
		Positive and negative values are color-coded with red and blue, respectively.
		From left to right: no temporal loss applied, $\mathcal{L}_{2,t}$ loss applied, 
		$\mathcal{L}_{D_t'}$, i.e., applied without advection, $\mathcal{L}_{D_t}$ 
		applied with advection (our full tempoGAN approach), and the ground-truth $y$.
		From left to right across the different versions, the derivatives become 
		less jagged and less noisy, as well as more structured and narrow.
		This means the temporal coherence is improved, esp. for the result
		from our algorithm ($\mathcal{L}_{D_t}$).}
	\label{fig:tempoStill}
\end{figure}
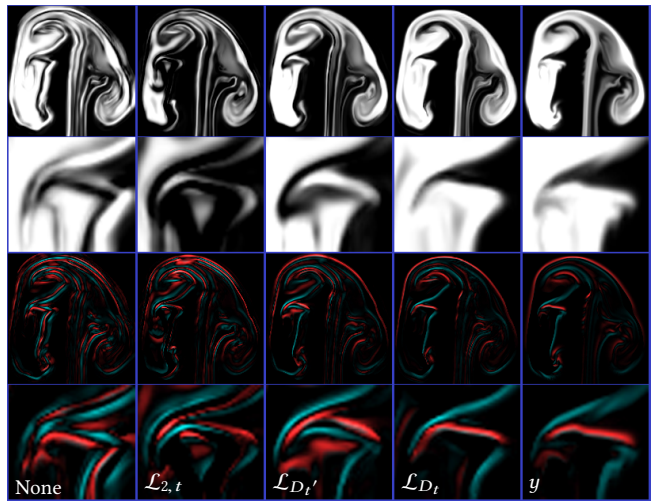

While this $\mathcal{L}_{2,t}$ based loss improves temporal coherence, our tests
show that its effect is relatively small. E.g., it can improve outlines, but leads to clearly unsatisfactory results,
which are best seen in the accompanying video. One side effect of this loss term
is that it can easily be minimized by simply reducing the values of $G(x)$. 
This is visible, e.g., in the second column of \myreffig{fig:tempoStill}, which contains noticeably less density
than the other versions and the ground truth. However, we do not want to drive the generator
towards darker outputs, but rather make it aware of how the data should change over time.

Instead of manually encoding the allowed temporal changes, 
we propose to use another discriminator $D_t$, that learns from the given data whose changes are admissible.
In this way, the original spatial discriminator, which we will denote as $D_s(x,G(x))$ from now on, guarantees that 
our generator learns to generate realistic details, while the new temporal discriminator $D_t$ mainly focuses on 
driving $G(x)$ towards solutions that match the temporal evolution of the
ground-truth $y$.

Specifically, $D_t$ takes three frames as input. We will denote such sets of three frames with a tilde in the following.
As real data for the discriminator, the set $\widetilde{Y}_{\mathcal{A}}$ contains three consecutive and advected frames, thus
$\widetilde{Y}_{\mathcal{A}} = \{ \mathcal{A}( y^{t-1}, v_x^{t-1} )$, $y^{t}$, $\mathcal{A}( y^{t+1}, -v_x^{t+1} )\}$. 
The generated data set contains correspondingly advected samples from the generator:
$\widetilde{G}_{\mathcal{A}}(${\footnotesize{$\widetilde{X}$}}$) = \{ \mathcal{A}( G(x^{t-1}), v_{x}^{t-1})$, $G(x^{t})$, $\mathcal{A}( G(x^{t+1}), -v_{x}^{t+1}) \}$. 

Similar to our spatial discriminator $D_s$, the temporal discriminator  $D_t$ is trained as a
binary classifier on the two sets of data:
\resizeEq{
\begin{split}
\mathcal{L}_{D_t}(D_t,G)= \myavg_m [-\log D_t\Big(\widetilde{Y}_{\mathcal{A}}\Big) ]
+ \myavg_n [-\log (1-D_t\Big(\widetilde{G}_{\mathcal{A}}\Big(\widetilde{X}\Big)\Big))]  \
\end{split}}{eq:tempod}
, where set $\widetilde{X}$ also contains three consecutive frames, i.e., $\widetilde{X} =\{x^{t-1},$ $ x^t, x^{t+1}\} $.
Note that unlike the spatial discriminator, $D_t$ is not a 
conditional discriminator. It does not ``see'' the conditional input $x$, and thus
$D_t$ is forced to make its judgment purely based on the given sequence.

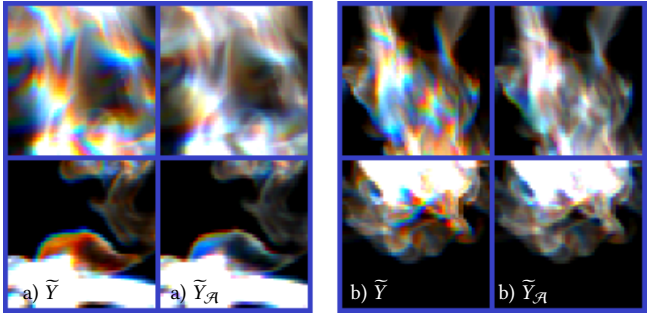
\begin{figure}[b]
	\centering 
	\begin{overpic}[width=\linewidth]{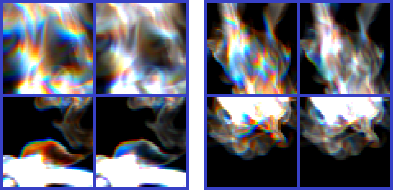}
		\put( 3.0,2.0){\small \color{black}{a) $\widetilde{Y}$}}
		\put(26.0,2){\small \color{black}{a) $\widetilde{Y}_{\mathcal{A}}$}}
		\put(54.0,2){\small \color{white}{b) $\widetilde{Y}$}}
		\put(77.0,2){\small \color{white}{b) $\widetilde{Y}_{\mathcal{A}}$}}
	\end{overpic}
	\caption{ These images highlight data alignment due to advection. 
		Three consecutive frames are encoded as R, G, B channels
		of a single image, thus, ideally a fully aligned image would only contain shades of grey.
		The two rows contain front and top views in the top and bottom row, respectively.
        We show two examples, a) and b). Each of them contains $\widetilde{Y}$ left, 
        and $\widetilde{Y}_{\mathcal{A}}$ right.
		The RGB channels are the three input frames, t-1, t, and t+1.
        Compared with $\widetilde{Y}$, $\widetilde{Y}_{\mathcal{A}}$ is significantly less saturated, 
		i.e., better aligned. }
	\label{fig:tempoData}
\end{figure}

In \myreffig{fig:tempoStill}, we show a comparison of the different loss variants for improving temporal
coherence. The first column is generated with only the spatial discriminator, i.e., 
provides a baseline for the improvements.
The second column shows the result using the $L_2$-based temporal loss $\mathcal{L}_{2,t}$ from \myrefeq{eq:l2VelDoubleSide}, 
while the fourth column shows the result using $D_t$ from \myrefeq{eq:tempod}. The last column is the ground-truth data $y$.
The first two rows show the generated density fields. While $\mathcal{L}_{2,t} $ reduces overall density content, the result 
with $D_t$ is clearly closer to the ground truth.
The bottom two rows show time derivatives of the densities for frames t and t+1. 
Again, the result from $D_t$ and the ground-truth $y$ match closely in terms of their time derivatives.
The large and jagged values of the first two rows indicate the undesirable temporal changes produced
by the regular GAN and the $\mathcal{L}_{2,t}$ loss.

In the third column of \myreffig{fig:tempoStill}, we show a simpler variant of 
our temporal discriminator. Here, we employ the discriminator without aligning the
set of inputs with advection operations, i.e.,
\resizeEq{
\begin{split}
\mathcal{L}_{D_t'}(D_t',G)=& \myavg_m [-\log D_t\Big(\widetilde{Y}\Big)] + \myavg_n 
[-\log (1-D_t(\widetilde{G}\Big(\widetilde{X}\Big) ))]
\end{split}}{}
with $\widetilde{Y} = \{y^{t-1}, y^t, y^{t+1} \}$ 
and $\widetilde{G}\Big(\widetilde{X}\Big) = \{G(x^{t-1}), G(x^{t}), G(x^{t+1})\}$.

\begin{figure*}[tb]
     \centering 
	\includegraphics[width=0.99\linewidth]{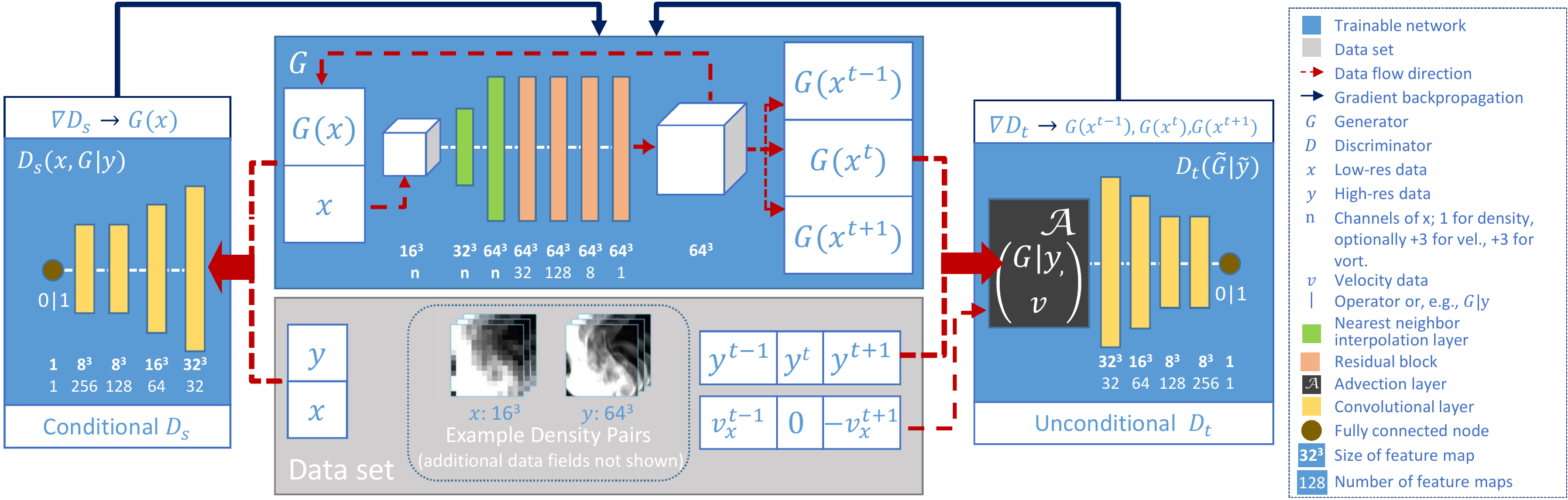} \\
     \caption{ Here an overview of our tempoGAN architecture is shown. The three neural networks (blue boxes)
     are trained in conjunction. The data flow between them is highlighted by the red and black arrows. \revision{Note that $x$ and $y$ denote fluid data that contains velocity and/or vorticity fields, as well as density depending on the chosen architecture (see \myrefsec{sec:inputs}).} }
     \label{fig:nnArchResnet}
\end{figure*}

This version improves results compared to $\mathcal{L}_{2,t}$, but does not 
reach the level of quality of $\mathcal{L}_{D_t}$, as can be seen in \myreffig{fig:tempoStill}.
Additionally, we found that  $\mathcal{L}_{D_t}$
often exhibits a faster convergence during the training process. This is an indication that the underlying
neural networks have difficulties aligning and comparing the data by themselves when using  $\mathcal{L}_{D_t'}$. 
This intuition is illustrated in \myreffig{fig:tempoData}, where
we show example content of the regular data sets $\widetilde{Y}$ and the advected version
$\widetilde{Y}_{\mathcal{A}}$ side by side. In this figure, the three chronological frames
are visualized as red, green, and blue channels of the images. Thus, a pure gray-scale image
would mean perfect alignment, while increasing visibility of individual colors indicates un-aligned
features in the data. \myreffig{fig:tempoData} shows
that, although not perfect, the advected one leads to clear improvements in terms of aligning the features of the data sets,
despite only using the approximated coarse velocity fields $v_x$. Our experiments show
that this alignment successfully improves the backpropagated gradients such that the
generator learns to produce more coherent outputs.
However, when no flow fields are available, $\mathcal{L}_{D_t'}$ still represents a better choice than the simpler $\mathcal{L}_{2,t}$ version.
We see this as another indicator of the power of adversarial training models. It seems
to be preferable to let a neural network learn and judge the specifics of a data set, instead
of manually specifying metrics, as we have demonstrated for 
data sets of fluid flow motions above.

It is worth pointing out that our formulation for $D_t$ in \myrefeq{eq:tempod} means
that the advection step is an inherent part of the generator training process. While
$v_x$ can be pre-computed, it needs to be applied to the outputs of the generator during
training. This in turn means that the advection needs to be tightly integrated into the training loop.
The results discussed in the previous paragraph indicate that if this is done correctly, the loss gradients of the
temporal discriminator are successfully passed through the advection steps to give the generator
feedback such that it can improve its results. In the general case, advection is a non-linear function, the discrete
approximation for which we have abbreviated with $\mathcal{A}(y^{t}, v_y^{t})$ above. Given a known 
flow field $v_y$ and time step, we can linearize this equation to yield a matrix $M y = \mathcal{A}(y^{t}, v_y^{t}) = y^{t+1}$.
E.g., for a first order approximation, $M$ would encode the Euler-step lookup of source positions
and linear interpolation to compute the solution. While we have found first order scheme (i.e., semi-Lagrangian
advection) to work well, $M$ could likewise
encode higher-order methods for advection.

We have implemented this process as an advection layer in our network training, which computes
the advection coefficients, and performs the matrix multiplication such that the 
discriminator receives the correct sets of inputs. When training the generator, the same 
code is used, and the underlying NN framework can easily compute the necessary derivatives.
In this way, the generator actually receives three accumulated, and aligned gradients
from the three input frames that were passed to $D_t$.

\subsection{Full Algorithm}
\label{sec:totalloss}
While the previous sections have explained the different parts of our final loss function,
we summarize and discuss the combined loss in the following section.
We will refer to our full algorithm as {\em tempoGAN}.
The resulting optimization problem that is solved with NN training consists
of three coupled non-linear sub-problems: the generator, the conditional spatial discriminator,
and the un-con\-di\-tio\-nal temporal discriminator. The generator has to effectively minimize both discriminator losses, additional feature space constraints,
and a $L_1$ regularization term. Thus, the loss functions can be summarized as:
\resizeEq{
\begin{split}
\mathcal{L}_{D_t}(D_t,G)= & -\myavg_m [\log D_t(\widetilde{Y}_{\mathcal{A}})] -
 \myavg_n [\log \Big(1-D_t\Big(\widetilde{G}_{\mathcal{A}}\Big(\widetilde{X}\Big)\Big)\Big)] \\
\mathcal{L}_{D_s}(D_s,G)=& -\myavg_m [\log D_s(x, y)] - \myavg_n [\log (1-D_s(x, G(x)))] \\
\mathcal{L}_{G}(D_s,D_t,G)=& -\myavg_n [\log D_s(x, G(x)) ]
- \myavg_n [\log D_t\Big(\widetilde{G}_{\mathcal{A}}\Big(\widetilde{X}\Big)\Big)] \\
& + \myavg _{n,j} \lambda _{f}^{j} \left \| F^{j}(G(x))-F^{j}({y}) \right \|_{2}^{2} + \lambda _{L_1}  \myavg_n \left \| G(x)-{y}\right \|_{1}
\end{split}}{eq:totalLoss}

Our generator has to effectively compete against two powerful adversaries,
who, along the lines of  "the enemy of my enemy is my friend", 
implicitly cooperate to expose the results of the generator. E.g., we have 
performed tests without $D_s$, only using $D_t$, and the resulting generator
outputs were smooth in time, but clearly less detailed than when using both discriminators.

Among the loss terms of the generator, the $L_1$ term has a relatively minor role to stabilize
the training by keeping the averaged output close to the target. However, due to the complex
optimization problem, it is nonetheless helpful for successful training runs.
The feature space loss, on the other hand, directly influences the generated features.
In the adversarial setting the discriminator most likely learns distinct features 
that only arise for the ground truth (positive features), or those that make it easy 
to identify generated versions, i.e., negative features that are only produced by the generator.
Thus, while training, the generator will receive gradients to make it produce more
features of the targets from $F(y)$, while the gradients from $F(G(x))$ will penalize
the generation of recognizable negative features. 

While positive values for $\lambda_f$ reinforce this behavior, it is less clear
why negative values can lead to even better results in certain cases. 
\revision{Our explanation
for this behavior is that the negative weights drive the generator towards distinct features that
have to adhere to the positive and negative features detected by the discriminator,
as explained above in \myrefsec{sec:featureloss}, but at the same time differ from the average features in $y$.}
Thus, the generator cannot simply create different or no features,
as the discriminator would easily detect this. Instead it needs to develop features
that are like the ones present in the outputs $y$, but
don't correspond to the average features in $F(y)$,
which, e.g., leads to the fine detailed outputs shown in \myreffig{fig:2dlayerlossexample}. 

\begin{algorithm}
	\revision{
		\caption{tempoGAN training algorithm}\label{alg:tempoGANalg}		
		\begin{algorithmic}[1]   \small    
			\For{number of training steps}
			\For{$k_{D_{s}}$}
			\State Compute data-augmented mini batch $x, y$
			\State Update $D_{s}$ with $\nabla_{D_{s}}[\mathcal{L}_{D_s}(D_s,G)]$
			\EndFor
			\For{$k_{D_{t}}$}
			\State Compute data-augmented mini batch $\widetilde{X}, \widetilde{Y}$
			\State Compute advected frames $\widetilde{Y}_{\mathcal{A}}$ and $\widetilde{G}_{\mathcal{A}}\Big(\widetilde{X}\Big)$
			\State Update $D_{t}$ with $\nabla_{D_{t}}[\mathcal{L}_{D_t}(D_t,G)]$
			\EndFor
			\For{$k_{G}$}
			\State Compute data-augmented mini batch $x, y, \widetilde{X}$
			\State Compute advected frames $\widetilde{G}_{\mathcal{A}}\Big(\widetilde{X}\Big)$
			\State Update $G$ with $\nabla_{G}[\mathcal{L}_{G}(D_s,D_t,G)]$
			\EndFor
			\EndFor
		\end{algorithmic}
	}
\end{algorithm}

\section{Architecture and Training Data}
\label{sec:nnarch}
While our loss function theoretically works with any realization of $G, D_s$ and $D_t$, their
specifics naturally have significant impact on performance and the quality of the generated outputs.
A variety of network architectures has been proposed
for training generative models \cite{goodfellow2014generative,RadfordMC15,berthelot2017began}, and
in the following, we will focus on pure convolutional networks for the generator, i.e., networks without any
fully connected layers. 
A fully convolutional network has the advantage that the trained network can
be applied to inputs of arbitrary sizes later on. 
We have experimented with a large variety of generator architectures, and while many 
simpler networks only yielded sub-optimal results,
we have achieved high quality results with generators based on the popular U-net \cite{ronneberger2015u, isola2016image}, 
as well as with residual networks ({\em res-nets}) \cite{lim2017enhanced}.
The U-net concatenates activations from earlier layers to later layers 
(so called {\em skip connections}) in order to allow the network to combine high- and low-level information, 
while the res-net
processes the data using by multiple {\em residual blocks}. Each of these residual blocks convolves the
inputs without changing their spatial size, and the result of two convolutional layers is added to the original 
signal as a ``residual'' correction. In the following, we will focus on the latter architecture, as it gave slightly 
sharper results in our tests. 

We found the discriminator architecture to be less crucial. 
As long as enough non-linearity is introduced over the course of several hidden layers,
and there are enough weights, changing the connectivity of the discriminator did not significantly influence
the generated outputs. Thus, in the following, we will always use discriminators with four convolutional layers
with leaky ReLU activations~\!\footnote{With a leaky tangent of 0.2 for the negative half space.}
followed by a fully connected layer to output the final score. 
\revision{As suggested by Odena et al. \shortcite{odena2016deconv}, we use the 
nearest-neighbor interpolation layers as the first two layers in our generator, instead of deconvolutional ones,
and in the discriminator networks, the kernel size is divisible by the corresponding stride.
An overview of the architecture of our neural 
networks is shown in \myreffig{fig:nnArchResnet}, while their details, such as layer configuration and activation 
functions, can be found in \myrefapp{app:nnarch}.}
\subsection{Data Generation and Training} \label{sec:datagen}

\revision{We use a randomized smoke simulation setup to generate the desired number of
training samples. For this  
we employ a standard fluids solve \cite{Stam1999} with MacCormack advection and MiC-preconditioned CG solver.}
We typically generate around 20 simulations,
with 120 frames of output per simulation. For each of these, we randomly 
initialize a certain number of smoke inflow regions, another set of velocity inflows,
and a randomized buoyancy force.  As inputs $x$,
we use a down-sampled version of the simulation data sets, typically by a factor of 4,
while the full resolution data is used as ground truth $y$. 
\revision{Note that this
setup is inherently {\em multi-modal}: for a single low resolution configuration,
an infinitely large number of correct high resolution exists. We do not explicitly
sample the high resolution solution space, but the down-sampling
in conjunction with data augmentation lead to ambiguous low- and high-resolution
pairs of input data.}
To prevent a large number of primarily empty samples, we discard inputs with
average smoke density of less than 0.02.
Details of the parametrization can be found in \myrefapp{app:data}, and visualizations of the training data
sets can be found in the supplemental video.
In 
addition, we show examples generated from a two-dimensional rising smoke 
simulation with a different simulation setup than the one used for generating
the training data. It is, e.g., used in \myreffig{fig:2dlayerlossexample}.

We use the same modalities for all training runs:
we employ the commonly used ADAM optimizer~\!\footnote{Parameterized with $\beta=0.5$.} 
with an initial learning rate of $2 \cdot 10^{-4}$ 
that decays  to 1/20th for second half of the training iterations. 
\revision{All parameters were determined experimentally, details are given in \myrefapp{app:data}.
The number of training iterations is typically on the order of 10k. 
We use 20\% of the data for testing and the remaining 80\% for training.
Our networks did not require any additional regularization such 
as dropout or weight decay. 
\revision{The training procedure is summarized again in \myrefalg{alg:tempoGANalg}. 
Due to the typically limited amount of GPU memory, especially for 3D data sets, we perform multiple training steps
for each of the components. 
Detail are listed in \myrefapp{app:data}.
In \myrefalg{alg:tempoGANalg}, we use $k_{D_{s}}$, $k_{D_{t}}$, and $k_{G}$ to denote the number training iterations for $D_{s}$, $D_{t}$, and $G$, respectively.
}\\

While the coupled non-linear optimization can yield different results
even for runs with the same parameters due to the non-deterministic nature
of parallelized operations, we found the results to be stable in terms of quality.
In particular, we did not find it necessary to change the weights of the different
discriminator loss terms. However, if desired, $\lambda_{f}$  can be used to influence
the learned details as described above.}
For training and running the trained networks, we use 
Nvidia GeForce GTX 1080 Ti GPUs (each with 11GB Ram) and Intel Core i7-6850K CPUs,
while we used the {\em tensorflow} and {\em mantaflow} software frameworks for
deep learning and fluid simulation implementations, respectively.

\subsection{  Input Fields }
\label{sec:inputs}

\begin{figure}[tb]
    \begin{overpic}[width=0.74 \linewidth]{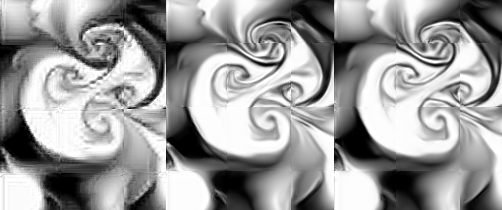}
		\put( 2,3){\small \color{black}{a)}}
		\put(35,3){\small \color{black}{b)}} 
		\put(68,3){\small \color{black}{c)}}  
	\end{overpic}
    \begin{overpic}[width=0.2466 \linewidth]{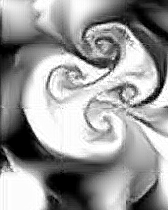}
		\put(6 ,6){\small \color{black}{d)}} 
	\end{overpic}
	\caption{ An illustration of different training results after 40k iterations with different input
		fields: a) $\dens$, b) $\dens+\vel$, c) $\dens+\vel+\vort$,
		all with similar network sizes. Version d) with only $\dens$ has 2x the number of weights.
		The seams in the images show the size of the training patches. 
		Supplemental physical fields lead to clear improvements in b) and c), that even additional 
		weights cannot compensate for.}
	\label{fig:densVelVort}
\end{figure}
\begin{figure}[tb]
	\centering 
	\begin{overpic}[width=0.49\linewidth]{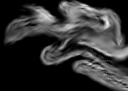}
		\put( 3,3){\small \color{white}{a)}}
	\end{overpic}
	\begin{overpic}[width=0.49\linewidth]{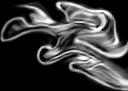}
		\put( 3,3){\small \color{white}{b)}}
	\end{overpic}
	\caption{ 
		An identical GAN network trained with the same set of input data
		While version a) did not use data augmentation, leading to blurry results with streak-like artifacts,
		version b), with data augmentation, produced sharp and detailed outputs.
	} \label{fig:dataAugComp}
\end{figure}

On first sight, it might seem redundant and unnecessary to input flow velocity $\vel$ and vorticity
$\vort$ in addition to the density $\dens$. After all, we are only interested in the
final output density, and many works on GANs exist, which demonstrate that
detailed images can be learned purely based on image content.

However, over the course of numerous training runs, we noticed that giving the networks additional 
information about the underlying physics significantly improves convergence and quality of the inferred results. 
An example is shown in \myreffig{fig:densVelVort}. Here, we show how the training evolves
for three networks with identical size, structure and parameters, the only difference being the
input fields. From left to right, the networks receive $(\dens)$, $(\dens,\vel)$, and $(\dens,\vel,\vort)$.
Note that these fields are only given to the 
generator, while the discriminator always only receives $(\dens)$ as input. 
The version with only density passed to the generator, $G(\dens)$, fails to reconstruct smooth and detailed outputs. 
Even after 40000 iterations, the results exhibit strong grid artifacts and lack detailed structures. 
In contrast, both versions with additional inputs start to yield higher quality outputs earlier during training.
While adding $\vel$ is crucial, the addition of $\vort$ 
only yields subtle improvements (most apparent at the top of the images in \myreffig{fig:densVelVort}),
which is why we will use $(\dens,\vel)$ to generate our final results below. 
\newRachel{ The full training run comparison is in our supplemental video.} 

We believe that the insight that auxiliary fields help improving training and inference quality is a surprising
and important one. The networks do not get any explicit guidance on how to use the additional information.
However, it clearly not only learns to use this information, but also benefits from having this
supporting information about the underlying physics processes. 
While larger networks can potentially alleviate the quality problems of
the den\-si\-ty-only version, as illustrated in \myreffig{fig:densVelVort} d), 
we believe it is highly preferable to instead construct and train smaller, 
physics-aware networks. This not only shortens training times and accelerates convergence, but also makes evaluating
the trained model more efficient in the long run. The availability of physical inputs turned out
to be a crucial addition in order to successfully realize high-dimensional GAN outputs for space-time data, 
which we will demonstrate in \myrefsec{sec:results}. 
\subsection{  Augmenting Physical Data}
\label{sec:dataaug}
Data augmentation turned out to be an important component
of our pipeline due to the high dimensionality of our data sets and
the large amount of memory they require. 
Without sufficient enough training data, the adversarial training yields undesirable results
due to overfitting.
While data augmentation is common practice for natural images \cite{dosovitskiy2016discriminative, krizhevsky2012imagenet},
we describe several aspects below that play a role for physical data sets.

The augmentation process allows us to train networks having millions of weights
with data sets that only contain a few hundred samples without overfitting. 
At the same time, we can ensure that the trained networks respect the
invariants of the underlying physical problems, which is crucial
for the complex space-time data sets of flow fields that we are considering.
E.g., we know from theory that solutions obey Galilean invariance, and we can make
sure our networks are aware of this property not by providing large data sets, but instead
by generating data with different inertial frames on the fly while training. 

In order to minimize the necessary size of the training set without deteriorating the result quality,
we generate modified data sets at training time. We focus on spatial transformations, which
take the form of  $\tilde{\vec{x}}(\vec{p}) = \vec{x}( A \vec{p} )$, where $\vec{p}$ is a spatial position, and $A$ denotes
an $4\times4$ matrix. For applying augmentation, we distinguish three types of components of a data set: \begin{itemize}
\item {\em passive}: these components can be transformed in a straight forward manner as described above. An example of passive
	components are the advected smoke fields $\rho$, shown in many of our examples.
\item {\em directional}: the content of these components needs to be transformed in conjunction with the augmentation. A good example
	is velocity, whose directions need to be adjusted for rotations and flips, i.e., $\tilde{\vec{v}}(\vec{p}) = A_{3\times3}\vec{v}( A \vec{p} )$, where 
	$A_{3\times3}$ is the upper left $3\times3$ matrix of $A$.
\item {\em derived}: finally, derived components would be invalid after applying augmentation, and thus need to be re-computed. 
	\revision{A good example are physical quantities such as vorticity,
	which contain mixed derivatives that cannot be easily transformed into a new frame of reference. 
	However, these quantities typically can be calculated anew from other fields after augmentation.}
\end{itemize}

If the data set contains quantities that cannot be computed from other augmented fields, this unfortunately
means that augmentation cannot be applied easily. However, we believe that a large class of typical physics data
sets can in practice be augmented as described here.

\begin{figure*}[tb]
	\centering 
	\begin{overpic}[width=0.33\linewidth]{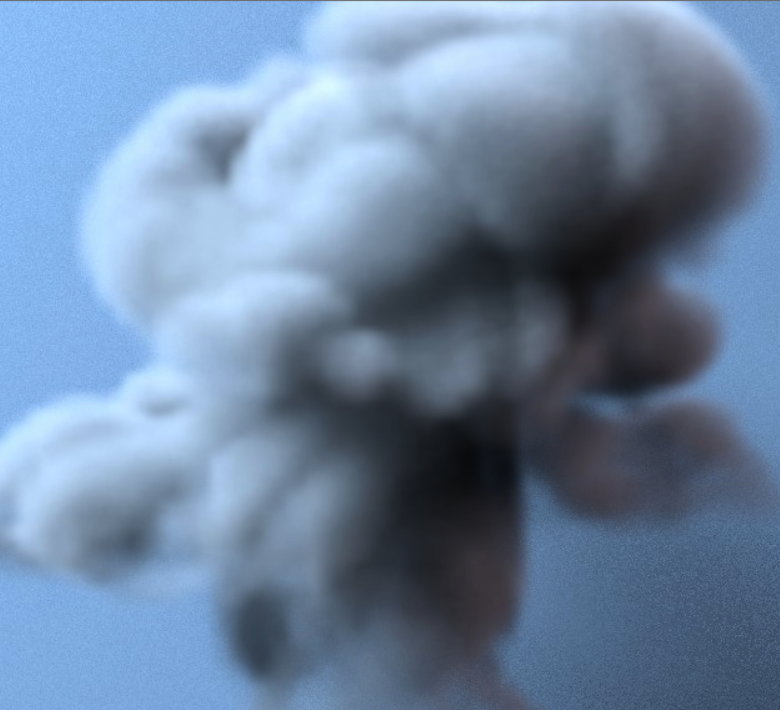}
		\put( 3,3){\small \color{blue1}{a)}}
	\end{overpic}
	\begin{overpic}[width=0.33\linewidth]{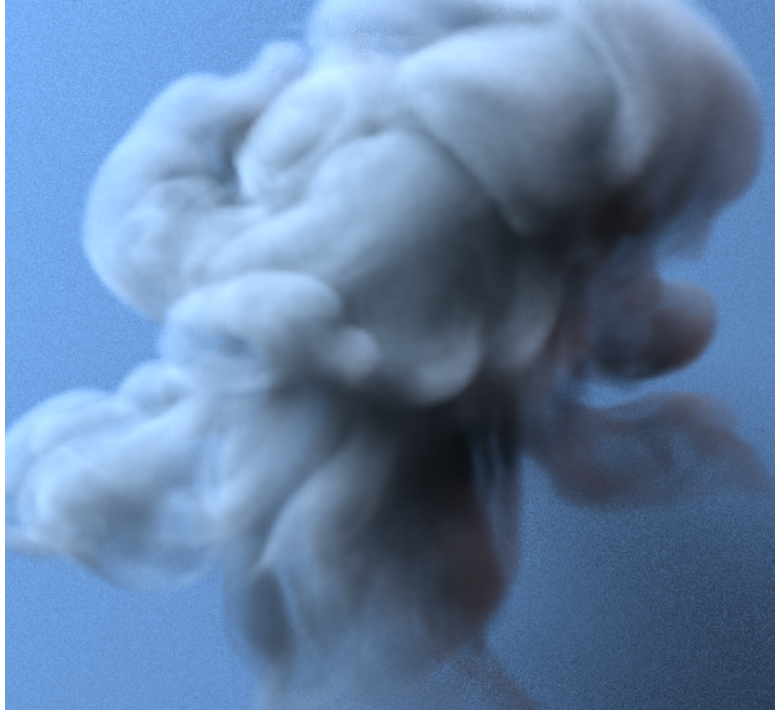}
		\put( 3,3){\small \color{blue1}{b)}}
	\end{overpic}
	\begin{overpic}[width=0.33\linewidth]{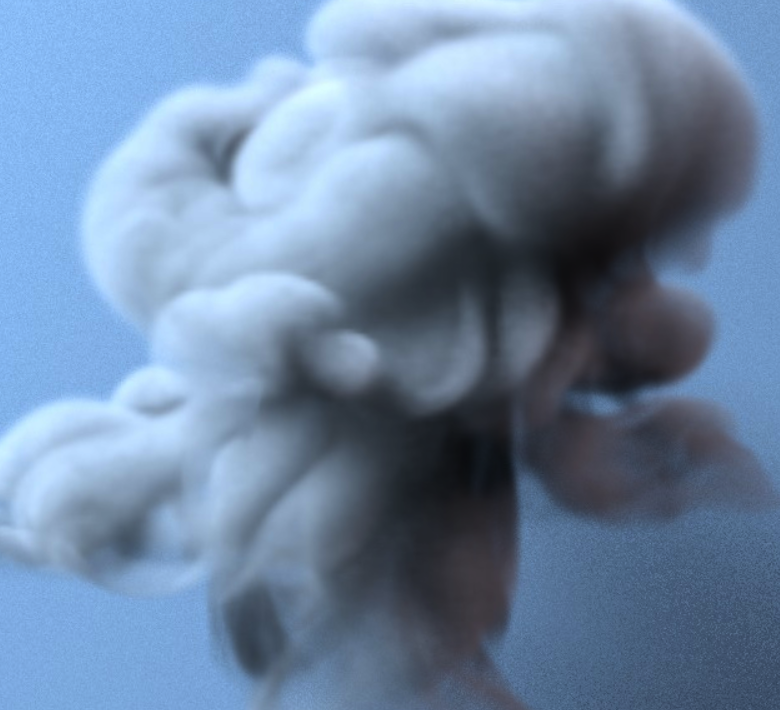}
		\put( 3,3){\small \color{blue1}{c)}}
	\end{overpic} 
	\caption{ These images show our algorithm applied to a 3D volume. F.l.t.r.: 
		\revision{a). a coarse input volume (down-sampled from the reference c, rendered with cubic up-sampling),}
		b). our result, and c). the high resolution reference. As in 2D, our trained model generates
		sharp features and detailed sheets that are at least on par with the reference. }
	\label{fig:plume3d}
\end{figure*}

For matrix $A$, we consider affine transformation matrices that contain combinations of randomized translations, uniform scaling, reflections,
and rotations. Here, only those transformations are allowed that do not violate the physical model
for the data set. \revision{While shearing and non-uniform scaling could easily be added, they violate the NS momentum equation 
and thus should not be used for flow data.}
We have used values in the range $[0.85, 1.15]$ for scaling, and rotations by $[-90, 90]$
degrees.
We typically do not load derived components into memory
for training, as they are re-computed after augmentation. Thus, they are computed on the fly for a training batch
and discarded afterwards.

The outputs of our simulations typically have significantly larger size than 
the input tiles that our networks receive.
In this way, we have many choices for choosing offsets, in order
to train the networks for shift invariance. This also aligns with our
goal to train a network that will later on work for arbitrarily sized inputs.
We found it important to take special care at spatial boundaries of the tiles. 
While data could be extended by Dirichlet or periodic boundary conditions, it is important that
the data set boundaries after augmentation do not lie outside the original data set. We enforce this 
by choosing suitable translations after applying the other transformations. This ensures that all data sets
contain only valid content, and the network does not learn from potentially unphysical or unrepresentative 
data near boundaries.
We also do not augment the time axis in the same way as the spatial axes. We found that the spatial transformations above applied to
velocity fields give enough variance in terms of temporal changes.
\begin{figure}[tb]
	\centering \includegraphics[width=\linewidth]{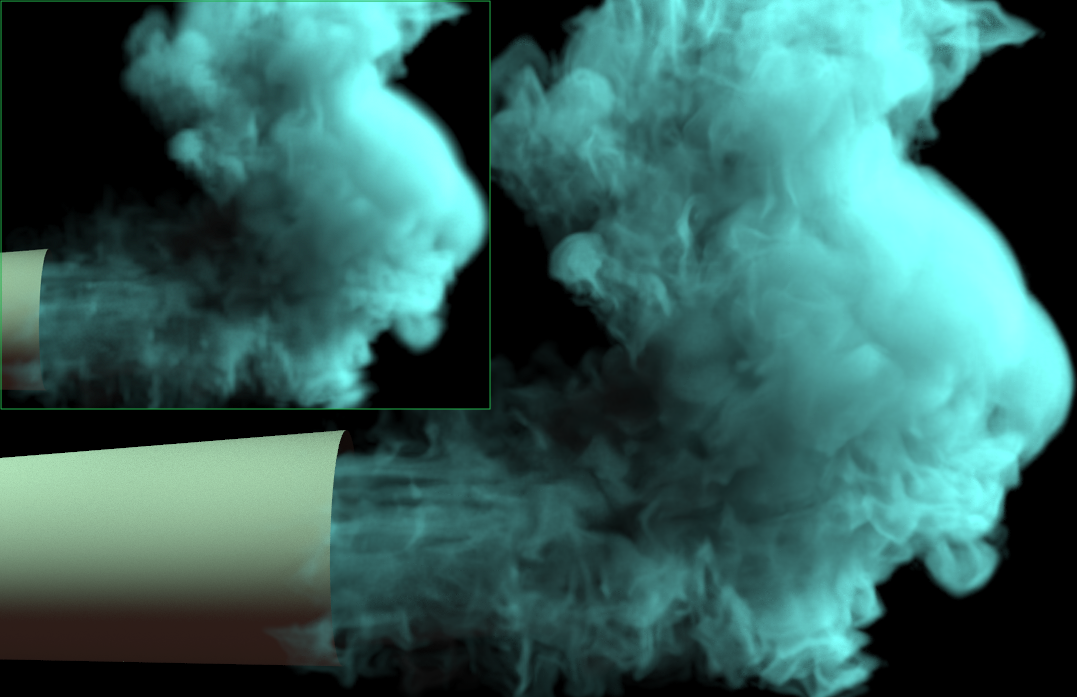}\\     
	\caption{ We apply our algorithm to a horizontal jet of smoke in this example. The inset
		shows the coarse input (rendered with cubic up-sampling), and the result
		of our algorithm. The diffuse streaks caused by procedural turbulence in the input
		(esp. near the inflow) are turned into detailed wisps of smoke by our algorithm. }
	\label{fig:green}
\end{figure}
\begin{figure}[tb]
	\centering \includegraphics[width=\linewidth]{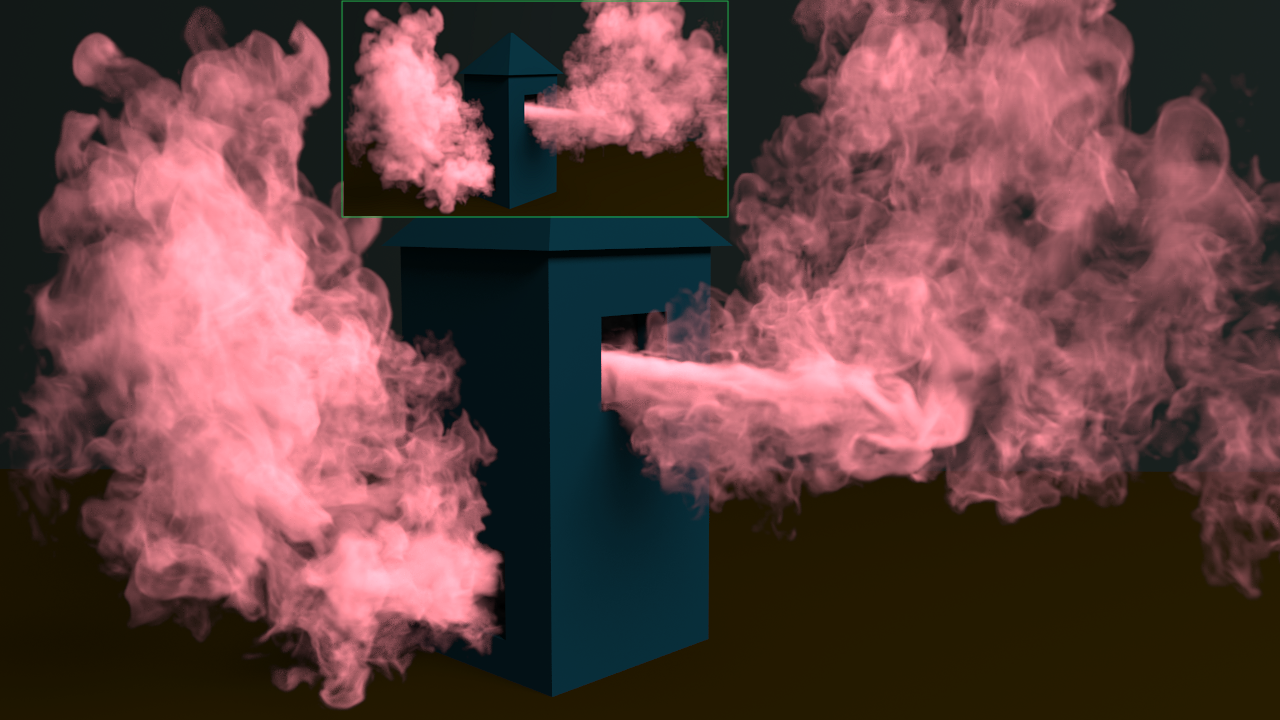}\\   
	\caption{Our algorithm generated a high-resolution volume around an obstacle
		with a final resolution of $1024 \times 720 \times 720$.
		The inset shows the input volume. This scene is also shown in \myreffig{fig:teaser}
		with a different visualization. }
	\label{fig:house}
\end{figure}
An example of the huge difference that data augmentation can make is shown in \myreffig{fig:dataAugComp}.
Here we compare two runs with the same amount of training data (160 frames of 
data), one with, the other one without data augmentation.
While training a GAN directly with this data produces blurry results,
the network converges to a final state with significantly sharper results with data augmentation. 
The possibility to successfully train networks with only a small amount
of training data is what makes it possible to train networks for 3D+time data, as 
we will demonstrate in \myrefsec{sec:results}.
\section{Results and Applications}
\label{sec:results}
In the following, we will apply our method discussed so far
to different data sets, and explore different application settings. Among others, we will discuss related topics
such as art direction, training convergence, and performance.~\footnote{We will make code and trained models available upon acceptance of our work.}
\subsection{3D Results}
We have primarily used the 2D rising plume example in the previous sections 
to ensure the different variants can be compared easily. 
In  \myreffig{fig:plume3d}, we demonstrate that these results directly extend
to 3D. \revision{We apply our method to a three-dimensional plume with resolution $64^3$, 
which in this case was generated by down-sampling a $256^3$ simulation such that we can compare our result to
this reference solution.
For this input data, the $256^3$ output produced by our tempoGAN exhibits small scale features that are at least
as detailed as the ground truth reference.} The temporal coherence is especially important
in this setting, which is best seen in the accompanying video.

We also apply our trained 3D model to two different inputs with higher resolutions. In both cases,
we use a regular simulation augmented with additional turbulence to generate an interesting
set of inputs for our method. A first scene with $150\times100\times100$ is shown in  \myreffig{fig:green}, 
where we generate a $600\times400\times400$
output with our method. The output closely resembles
the input volumes, but exhibits a large number of fine details.
\revision{Note that our networks were only trained with down-sampled inputs,
but our models generalize well to regular simulation inputs without re-sampling, as illustrated by this example.}

Our method also has no problems with obstacles in the flow, as shown in \myreffig{fig:house}.
This example has resolutions of $256 \times 180 \times 180$ and $1024 \times 720 \times 720$ for input and output volumes. 
The small-scale features closely adhere to the input flow around the obstacle. Although the obstacle is
completely filled with densities towards the end of the simulation, there are no leaking artifacts
as our method is applied independently to each input volume in the sequence.
When showing the low-resolution input, we always employ cubic up-sampling, in order
to not make the input look unnecessarily bad.

\subsection{\revision{Fine Tuning Results}}
\label{sec:artcontrol}
\revision{
GANs have a reputation for being particularly hard to influence and control,
and influencing the outcome of simulation results is an important topic for applications in computer graphics. 
In contrast to procedural methods,
regular GAN models typically lack intuitive control knobs to influence the generated results.}
While we primarily rely on traditional guiding techniques to control the low-resolution input,
our method offers different ways to adjust the details produced by our tempoGAN algorithm.
\begin{figure}[tb]
	\centering \begin{overpic}[width=0.99\linewidth]{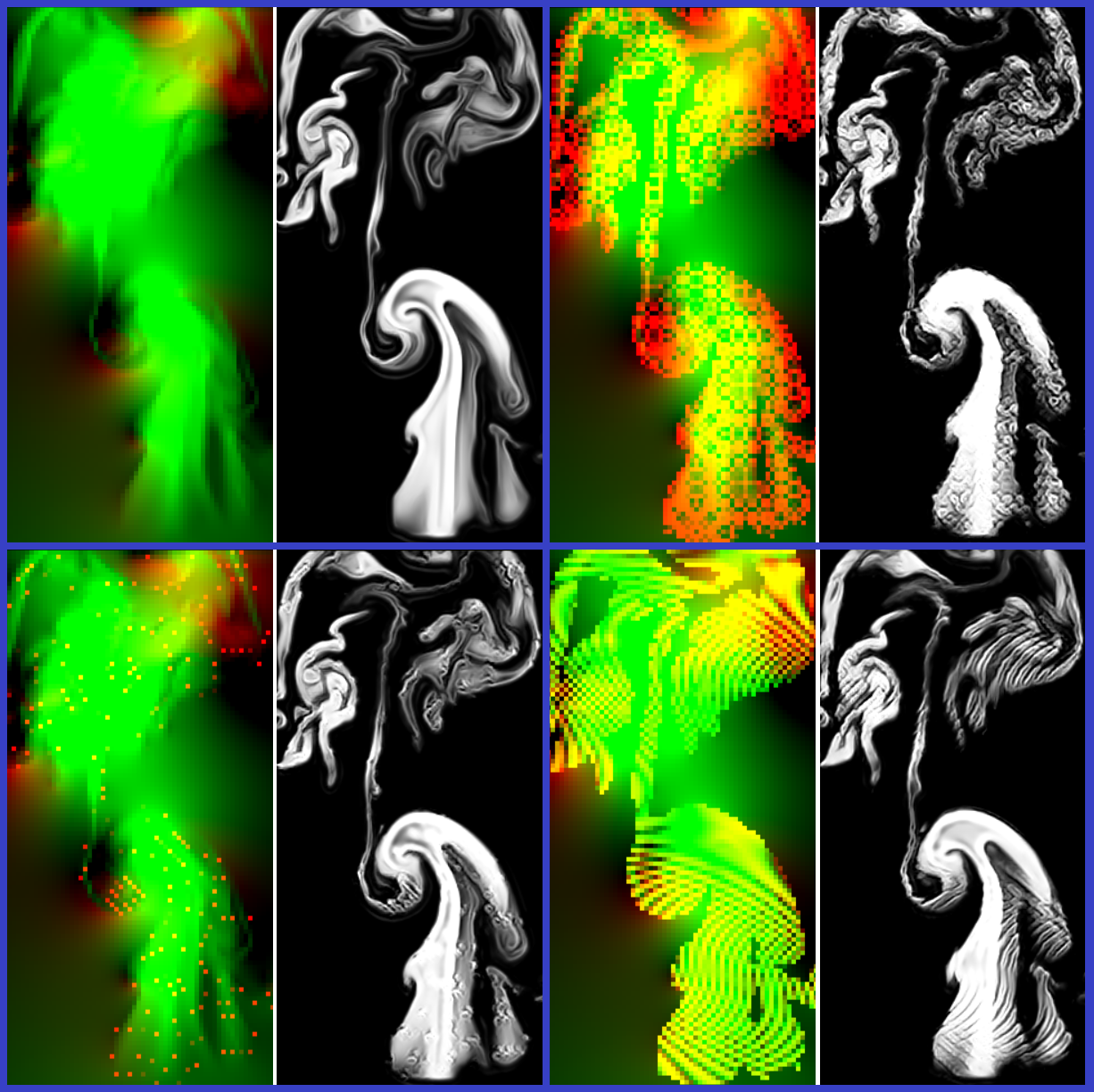}
		\put( 3,53){\small \color{white}{a)}}
		\put( 53,53){\small \color{white}{b)}}
		\put( 3,3){\small \color{white}{c)}}
		\put( 53,3){\small \color{white}{d)}}
	\end{overpic} 	
	\caption{The  red\&green images on the left of each pair represent the modified velocity inputs, while the corresponding result is shown on the right. For reference, pair a) shows the unmodified input velocity, and the regular output of our algorithm. }
	\label{fig:differentvelocity}
\end{figure}
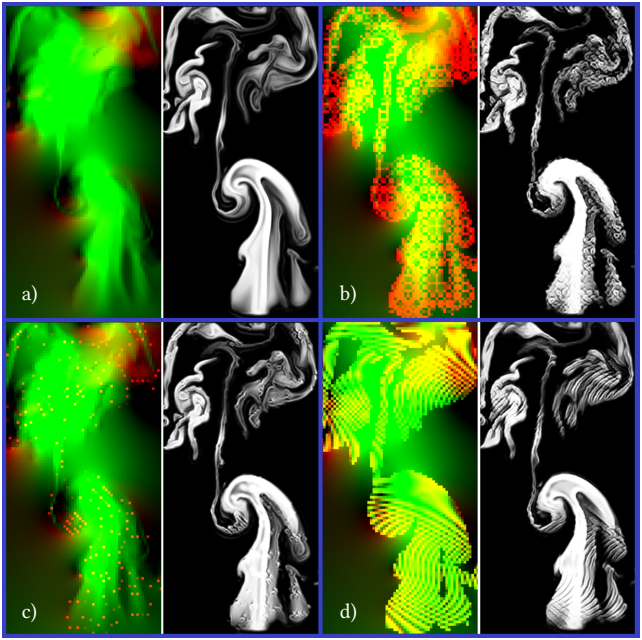

A first control knob for fine-tuning the results is to modify the data fields of the conditional
inputs. As described in \myrefsec{sec:inputs}, our generator receives the velocity in addition to the density, 
and it internally builds tight relationships between the two. 
We can use these entangled
inputs to control the features produced in the outputs. To achieve this, we modify the velocity components
passed to the generator with various procedural functions. \myreffig{fig:differentvelocity} shows the results of original input and several modified velocity
 examples and the resulting density configurations. 
\revision{We have also experimented with noise fields instead \cite{mirza2014conditional},
 but found that the trained networks completely ignored these fields. Instead, the strongly correlated velocity fields
naturally provide a much more meaningful input for our networks, and as a consequence provide means for influencing the
results.}

\begin{figure}[t]
    \centering 
	\begin{overpic}[width=0.99\linewidth]{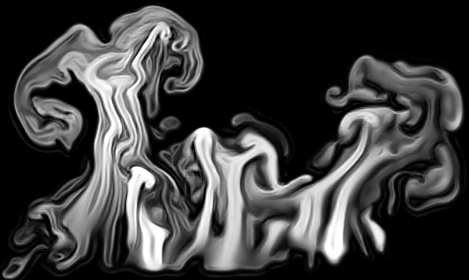}
	\end{overpic} 
    \caption{
	\revision{An illustration how the entangled inputs of density and velocity can be used to fine tune the results:
	on the left the velocities were scaled up by a factor of 2, while the right hand side was scaled by zero.
	The network has learned a relationship between detail and velocities, leading
	to reduced details in regions where the velocity was set to zero. }}
\label{fig:zerovelocity}
\end{figure}

\begin{figure}[t]
     \centering \begin{overpic}[width=0.49\linewidth]{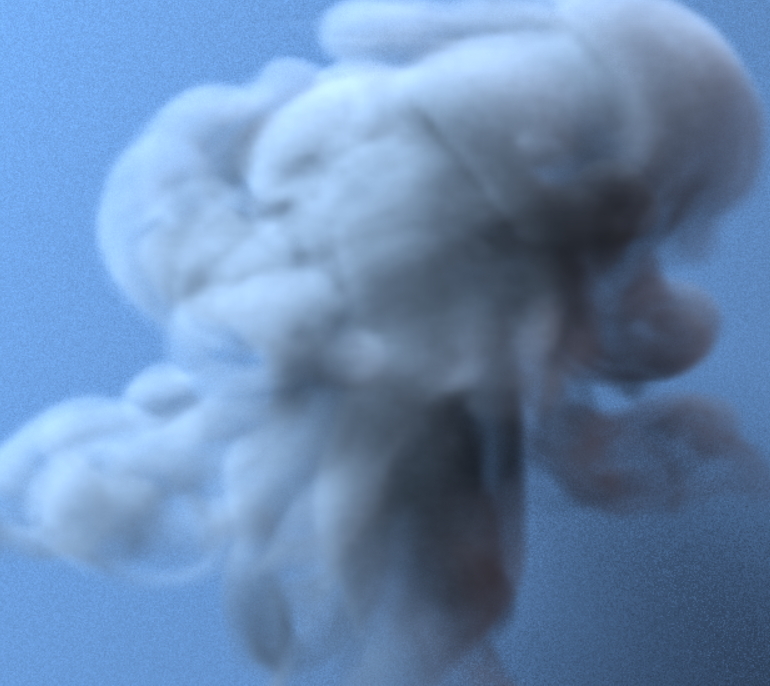}
     \put( 3,3){\small \color{blue1}{a)}}
     \end{overpic} 
     \begin{overpic}[width=0.49\linewidth]{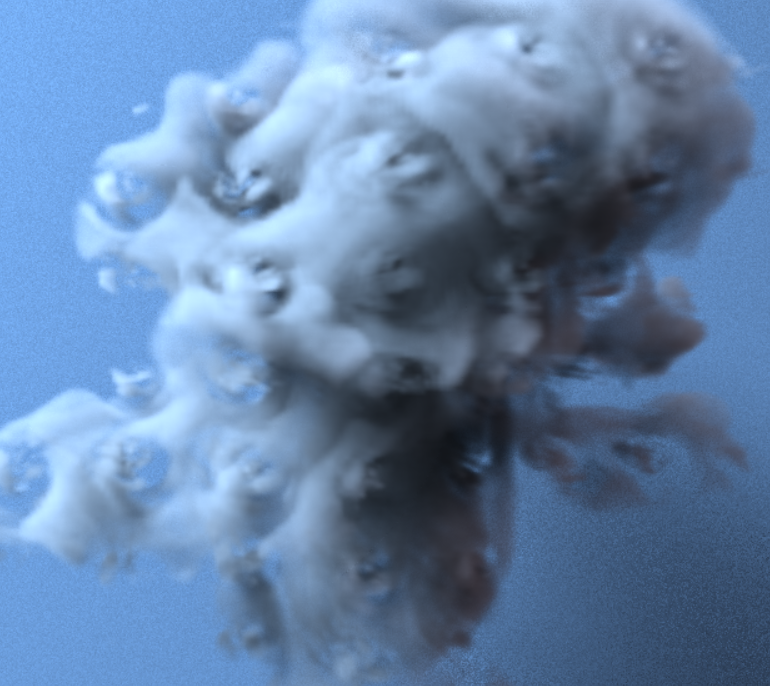}
     	\put( 3,3){\small \color{blue1}{b)}}
     \end{overpic} 
     \centering \begin{overpic}[width=0.49\linewidth]{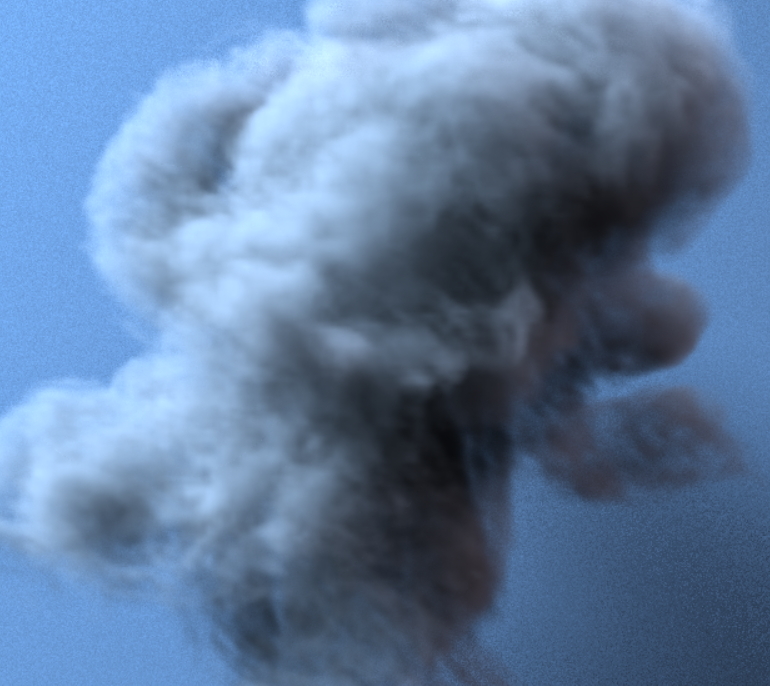}
     	\put( 3,3){\small \color{blue1}{c)}}
     \end{overpic} 
     \begin{overpic}[width=0.49\linewidth]{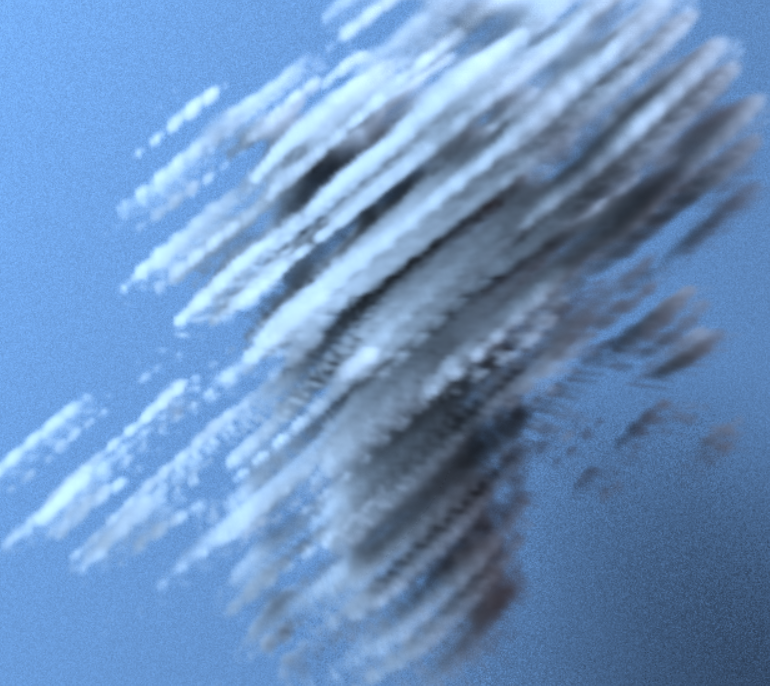}
     	\put( 3,3){\small \color{blue1}{d)}}
     \end{overpic} 
     \caption{a) is the result of tempoGAN with velocity set to zero.
     	The other
     three examples were generated with modified velocity inputs to achieve more stylized outputs.}
     \label{fig:3ddifferentvelocity}
\end{figure}

In addition, \myreffig{fig:zerovelocity} demonstrates that we can effectively suppress the generation
 of small scale details by setting all velocities to zero. Thus, the network learns a correlation between
 velocity magnitudes and amount of features. This is another indicator that the network learns to extract 
 meaningful relationships from the data, as we expect turbulence and small-scale details to primarily 
 form in regions with large velocities.
Three-dimensional data can similarly be controlled, as illustrated in \myreffig{fig:3ddifferentvelocity}. 

\begin{figure}[t]
	\centering \begin{overpic}[width=0.49\linewidth]{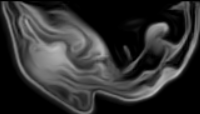}
		\put( 3,3){\small \color{white}{a)}}
	\end{overpic} 
	\centering \begin{overpic}[width=0.49\linewidth]{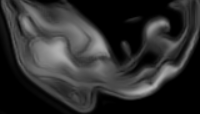}
		\put( 3,3){\small \color{white}{b)}}
	\end{overpic} 
	\caption{ A comparison of training runs with different feature loss weights: a) 
	$\lambda _{f}^{1,...,4}=-10^{-5}$ , 
	b) 
	$\lambda _{f}^{1,4}=1/3\cdot{10^{-4}}$, $\lambda _{f}^{2,3}=-1/3\cdot10^{-4}$. }
	\label{fig:styleLLCompare}
\end{figure}

\begin{figure}[t]
     \centering \begin{overpic}[width=0.32\linewidth]{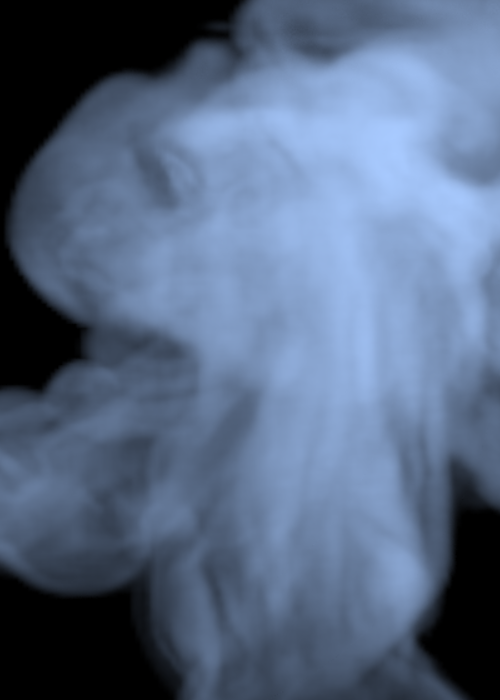}
     	     	\put( 3,3){\small \color{white}{a)}}
     \end{overpic} 
     \begin{overpic}[width=0.32\linewidth]{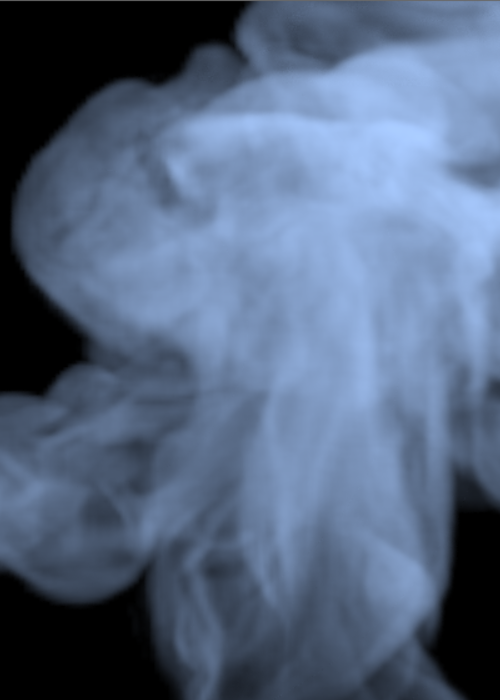}
     	     	\put( 3,3){\small \color{white}{b)}}
     \end{overpic} 
     \begin{overpic}[width=0.32\linewidth]{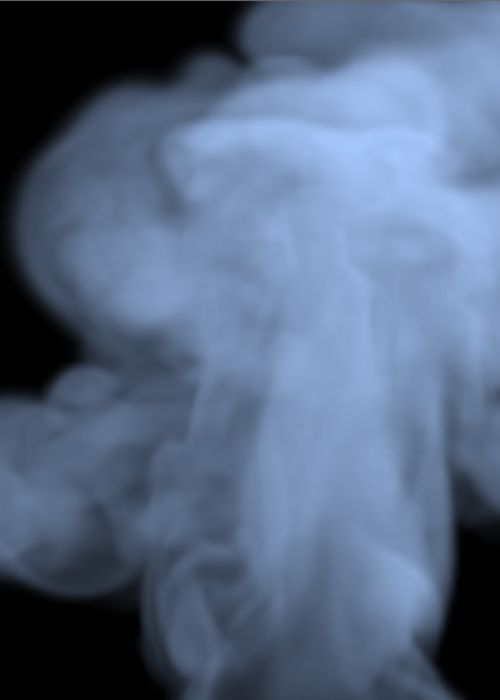}
     	     	\put( 3,3){\small \color{white}{c)}}
     \end{overpic} 
     \caption{ A comparison of training runs with different feature loss weights in 3D:
     a) with $\lambda _{f}^{1,...,4}=-1/3\cdot10^{-6}$ ,
     b) with $\lambda _{f}^{1}=1/3\cdot10^{-6}$, $\lambda _{f}^{2,3,4}=-1/3\cdot10^{-6}$. The latter yields a sharpened result.
          Image c) shows the high resolution reference. }
     \label{fig:3ddifferentstyle}
\end{figure}

In \myrefsec{sec:featureloss}, we discussed the influence of the $\lambda_{f}$ parameter for small scale features.
For situations where we might not have additional channels such as the velocity above, we can use $\lambda_{f}$
to globally let the network generate different features. However, as this only provides a uniform change that is encoded
in the trained network, the resulting differences are more subtle than those from the velocity modifications above. Examples
of different 2D and 3D outputs can be found in  \myreffig{fig:styleLLCompare} and  \myreffig{fig:3ddifferentstyle}, respectively.
\begin{figure}[t]
	\centering \begin{overpic}[width=0.49\linewidth]{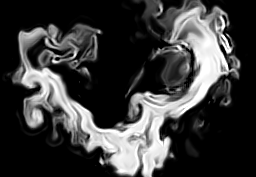}    
		\put( 3,3){\small \color{white}{a)}}
	\end{overpic} 
	\begin{overpic}[width=0.49\linewidth]{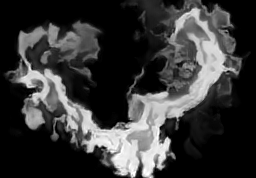}    
		\put( 3,3){\small \color{white}{b)}}
	\end{overpic} 
	\caption{ Our regular model a) and one trained with wavelet turbulence data
		b). In contrast to the model trained with real simulation data, the wavelet
		turbulence model produces flat regions with sharper swirls, mimicking
		the input data.
	}
	\label{fig:waveletTurbulence}
\end{figure}
\begin{figure}[t]
	\centering \begin{overpic}[width=0.49\linewidth]{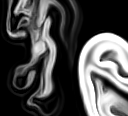}   
		\put( 3,3){\small \color{white}{a)}}
	\end{overpic}  
	\begin{overpic}[width=0.49\linewidth]{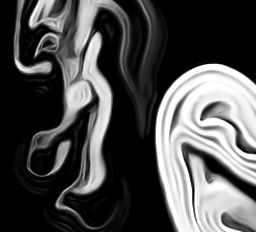}  
		\put( 3,3){\small \color{white}{b)}}
	\end{overpic}   
	\caption{a) is the network output after a single application. 
	b) is the network recursively applied to a) with a scaling factor of 2, resulting in a total increase of $8\times$.}
	\label{fig:recursive}
\end{figure}
\subsection{Additional Variants}

In order to verify that our network can not only work with two- or three-dimensional
data from a Navier-Stokes solver, we generated a more synthetic data set by applying
strong wavelet turbulence to a $4\times$ up-sampled input flow. We then trained our network
with down-sampled inputs, i.e., giving it the task to learn the output
of the wavelet turbulence algorithm. Note that a key difference here is that wavelet
turbulence normally requires a full high-resolution advection over time, while our method
infers high-resolution data sets purely based on low-resolution data from a single frame.
\begin{figure*}[t]
     \centering
		\includegraphics[width=0.99\linewidth, height = 3.3cm]{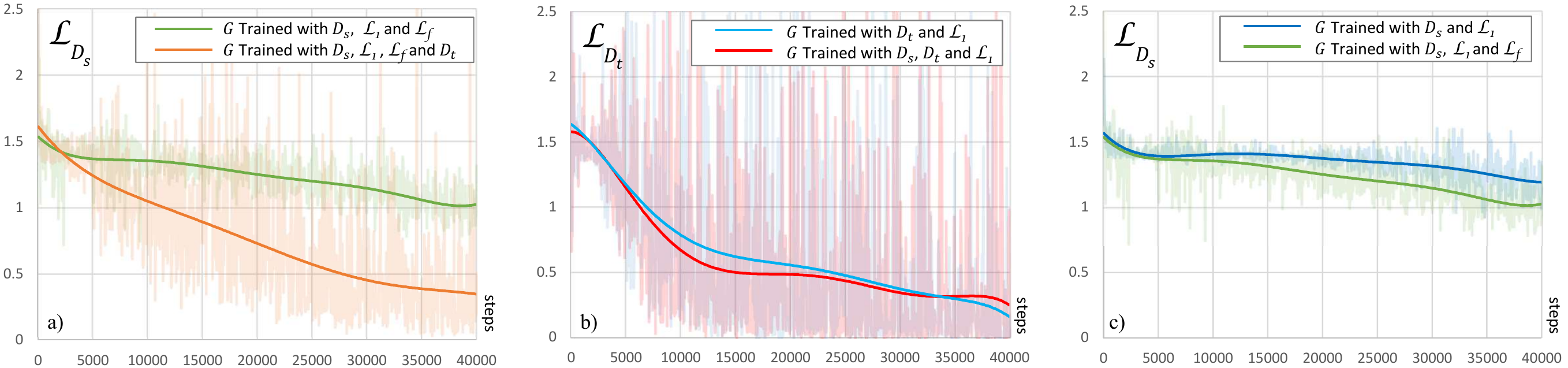}
     \caption{ Several discriminator loss functions over the course of the 40k training iterations.
     a) $D_{s}$ (spatial discriminator) loss is shown in green without $D_{t}$, and orange with $D_{t}$.
     b) Temporal discriminator loss in blue with only $D_{t}$, and in red for tempoGAN (i.e., with $D_{s}$, and feature loss).
     c) Spatial discriminator loss is shown in green with $\mathcal{L}_{f}$, and in dark blue without.
     For each graph, the dark lines show smoothed curves. The full data is shown in a lighter color in the background.
	}
     \label{fig:graph-all}
\end{figure*}

Our network successfully learns to generate structures similar 
to the wavelet turbulence outputs, shown in \myreffig{fig:waveletTurbulence}. However, this data set 
turned out to be more difficult to learn than the original fluid simulation inputs.
The training runs required two times more training data than the regular simulation runs,
and we used a feature loss of $\lambda _{f}^{1,...,4}=10^{-5}$. 
We assume that these more difficult training conditions are caused by the more 
chaotic nature of the procedural turbulence,
and the less reasonable correlations between density and velocity inputs. Note that
despite using more wavelet turbulence input data, it is still a comparatively small data set.

We additionally were curious how well our network works when it is applied
to a generated output, i.e., a recursive application. The result can be
found in \myreffig{fig:recursive}, where we applied our network 
to its own output for an additional $2\times$ upsampling. Thus, in total this led to an $8\times$ increase of resolution.
While the output is plausible, and clearly 
contains even more fine features such as thin sheets, there is a tendency
to amplify features generated during the first application.
\subsection{Training Progress}

With the training settings given in \myrefapp{app:data}, our training
runs typically converged to stable solutions of around $1/2$ for the discriminator outputs after sigmoid activation. 
While this by itself does not guarantee that a desirable solution was found, it at least indicates convergence
towards one of the available local minima.

\revision{However, it is interesting how the discriminator loss changes 
in the presence of the temporal discriminator.
\myreffig{fig:graph-all} shows several graphs of discriminator losses over the course of a full training run.
Note that we show the final loss outputs from \myrefeq{eq:disloss} and \myrefeq{eq:tempod} here.
A large value means the discriminator does ``worse'', i.e.,
it has more difficulty distinguishing real samples from the generated ones. 
Correspondingly, lower values mean it can separate them more successfully.
In \myreffig{fig:graph-all}a) it is visible that the spatial discriminator loss decreases when 
the temporal discriminator is introduced. 
Here the graph only shows the spatial discriminator loss, and the discriminator itself
is unchanged when the second discriminator is introduced. The training run corresponding to the
green line is trained with only a spatial discriminator, and for the orange line with both spatial and temporal discriminators.
Our interpretation
of the lower loss for the spatial discriminator network is that the existence of a temporal discriminator in the optimization 
prevents the generator from using
the part of the solution space with detailed, but flickering outputs. Hence, the generator is driven to find a solution from the temporally
coherent ones, and as a consequence has a harder time, which in turn makes the job easier for the spatial discriminator. This manifests
itself as a lower loss for the spatial discriminator, i.e. the lower orange curve in \myreffig{fig:graph-all}a).}

Conversely, the existence of a spatial discriminator does not noticeably influence
the temporal discriminator, as shown in \myreffig{fig:graph-all}b). This is also intuitive, as the spatial discriminator does not influence temporal changes. 
\revision{We found that a generator trained only with $D_t$ typically produces fewer details
than a generator trained with both. In conjunction, our tests indicate that the two discriminators 
successfully influence different aspects of the solution space,
as intended.}
Lastly, \myreffig{fig:graph-all}c) shows
that activating the negative feature loss from \myrefsec{sec:featureloss} makes the task for
the generator slightly harder, resulting in a lowered spatial discriminator loss. 
\subsection{Performance}

Training our two- and three-dimensional models
is relatively expensive. Our full 2D runs typically
take around 14 hours to complete (1 GPU), while the 3D runs
took ca. 9 days using two GPUs. However, in practice, the state of the
model after a quarter of this time is already indicative of
the final performance. The remainder of the time is typically
spent fine-tuning the network.

When using our trained network to generate high-resolution
outputs in 3D, the limited memory of current GPUs poses a constraint
on the volumes that can be processed at once, as the intermediate
layers with their feature maps can take up significant amounts of memory. 
However, this does not pose a problem for generating larger final
volumes, as we can subdivide the input volumes, and process them piece by piece.
We generate tiles with a size of $136^3$ on one GPU, with a corresponding input of size $34^3$. 
Our 8 convolutional layers with a receptive field of 16 cells mean that up to four 
cells of an input could be influenced by a boundary. In practice, we found 3 input cells
to be enough in terms of overlap. Generating a single $136^3$ output took ca. 2.2 seconds
on average. 
Thus, generating a $256^3$ volume from an $64^3$ input took 17.9s on average.
\revision{Comparing the performance of our model with high resolution simulations
is inherently difficult, due to the substantially different implementations and hardware platforms (CPU vs. GPU).
However, for the example of \myreffig{fig:green} we estimate that fluid simulation at the full resolution would
take ca. 31.5 minutes per frame of animation on average, while the evaluation of all volume
tiles with our evaluation pipeline took ca. 3.9 minutes.} 

\revision{
The cost for the trained model scales linearly with the number of cells in the volume,
and in contrast to all previous methods for increasing the resolution of flow simulations,
our method does not require any additional tracking information. It is also fully independent for all frames.
Thus, our method could ideally be applied on the fly before rendering a volume, after which the 
high resolution data could be discarded.
Additionally, due to GPU memory restrictions we currently evaluate our model in 
volumetric tiles with 3 cells of overlap for the input. This overlap can potentially be reduced further,
and become unnecessary when enough memory is available to process the full input volume at once.
}

\subsection{Limitations and Discussion}
\label{sec:limitations}

One limitation of our approach is that the network encodes a fixed
resolution difference for the generated details. While the initial up-sampling
layers can be stripped, and the network could thus be applied to inputs of any size,
it will be interesting to explore different up-sampling factors beyond the factor
of four which we have used throughout.
\revision{With our current implementation, our method can also be slower than, e.g., calculating
the advection for a high resolution grid. However, a high-res advection would typically not lead to
different dynamics than those contained in the input flow, and require a sequential solve
for the whole animation sequence.}
Our networks have so far also focused
on buoyant smoke clouds. While obstacle interactions worked in our tests,
we assume that networks trained for larger data sets and with other types of interactions
could yield even better results.

Our three-dimensional networks needed a long time to train, circa nine days
for our final model. Luckily, this is a one-time cost, and the network can be flexibly reused
afterwards. However, if the synthesized small-scale features need to be fine-tuned,
which we luckily did not find necessary for our work, the long runtimes could make this
a difficult process. 
The feature loss weights clearly also are data dependent, e.g., we used
different settings for simulation and wavelet turbulence data.
Here, it will be an interesting direction for future work to 
give the network additional inputs for fine tuning the results beyond the velocity
modifications which discussed in \myrefsec{sec:artcontrol}.

\section{Conclusions}
\label{sec:conclusions}
We have realized a first conditional GAN approach for four-di\-men\-sional 
data sets, and
we have demonstrated that it is possible to train generators that preserve temporal coherence
using our novel time discriminator.
The network architecture of this temporal discriminator, which ensures
that the generator receives gradient information even for complex transport processes, makes it 
possible to robustly train networks for temporal evolutions. We have shown
that this discriminator improves the generation of stable details as well as the learning process itself. 
At the same time, our fully convolutional networks can be applied to inputs of arbitrary size, and 
our approach  provides basic means for art direction of the generated outputs.
We also found it very promising to see that our CNNs are able to benefit from coherent, physical 
information even in complex 3D settings, which led to reduced network sizes.

Overall, we believe that our contributions yield a robust and very general method 
for generative models of physics problems, and for super-resolution flows in particular.
It will be highly interesting as future work to apply our tempoGAN to other physical problem settings,
or even to non-physical data such as video streams.

\begin{acks}
This work was funded by the ERC Starting Grant {\em realFlow} (StG-2015-637014). We would like to thank Wei He for helping with making the videos, and all members of the graphics labs of TUM, IST Austria and ETH Zurich for the thorough discussions.
\end{acks}
\bibliography{fluids} 


\begin{thebibliography}{58}


\ifx \showCODEN    \undefined \def \showCODEN     #1{\unskip}     \fi
\ifx \showDOI      \undefined \def \showDOI       #1{#1}\fi
\ifx \showISBNx    \undefined \def \showISBNx     #1{\unskip}     \fi
\ifx \showISBNxiii \undefined \def \showISBNxiii  #1{\unskip}     \fi
\ifx \showISSN     \undefined \def \showISSN      #1{\unskip}     \fi
\ifx \showLCCN     \undefined \def \showLCCN      #1{\unskip}     \fi
\ifx \shownote     \undefined \def \shownote      #1{#1}          \fi
\ifx \showarticletitle \undefined \def \showarticletitle #1{#1}   \fi
\ifx \showURL      \undefined \def \showURL       {\relax}        \fi
\providecommand\bibfield[2]{#2}
\providecommand\bibinfo[2]{#2}
\providecommand\natexlab[1]{#1}
\providecommand\showeprint[2][]{arXiv:#2}

\bibitem[\protect\citeauthoryear{Arjovsky, Chintala, and Bottou}{Arjovsky
  et~al\mbox{.}}{2017}]%
        {arjovsky2017wasserstein}
\bibfield{author}{\bibinfo{person}{Martin Arjovsky}, \bibinfo{person}{Soumith
  Chintala}, {and} \bibinfo{person}{L{\'e}on Bottou}.}
  \bibinfo{year}{2017}\natexlab{}.
\newblock \showarticletitle{{Wasserstein GAN}}.
\newblock \bibinfo{journal}{{\em arXiv:1701.07875\/}} (\bibinfo{year}{2017}).
\newblock


\bibitem[\protect\citeauthoryear{Bako, Vogels, McWilliams, Meyer, Nov{\'a}K,
  Harvill, Sen, Derose, and Rousselle}{Bako et~al\mbox{.}}{2017}]%
        {bako2017kernel}
\bibfield{author}{\bibinfo{person}{Steve Bako}, \bibinfo{person}{Thijs Vogels},
  \bibinfo{person}{Brian McWilliams}, \bibinfo{person}{Mark Meyer},
  \bibinfo{person}{Jan Nov{\'a}K}, \bibinfo{person}{Alex Harvill},
  \bibinfo{person}{Pradeep Sen}, \bibinfo{person}{Tony Derose}, {and}
  \bibinfo{person}{Fabrice Rousselle}.} \bibinfo{year}{2017}\natexlab{}.
\newblock \showarticletitle{Kernel-predicting convolutional networks for
  denoising Monte Carlo renderings}.
\newblock \bibinfo{journal}{{\em ACM Transactions on Graphics (TOG)\/}}
  \bibinfo{volume}{36}, \bibinfo{number}{4} (\bibinfo{year}{2017}),
  \bibinfo{pages}{97}.
\newblock


\bibitem[\protect\citeauthoryear{Berthelot, Schumm, and Metz}{Berthelot
  et~al\mbox{.}}{2017}]%
        {berthelot2017began}
\bibfield{author}{\bibinfo{person}{David Berthelot}, \bibinfo{person}{Tom
  Schumm}, {and} \bibinfo{person}{Luke Metz}.} \bibinfo{year}{2017}\natexlab{}.
\newblock \showarticletitle{{BeGAN: Boundary equilibrium generative adversarial
  networks}}.
\newblock \bibinfo{journal}{{\em arXiv:1703.10717\/}} (\bibinfo{year}{2017}).
\newblock


\bibitem[\protect\citeauthoryear{Bhattacharjee and Das}{Bhattacharjee and
  Das}{2017}]%
        {bhattacharjee2017temporal}
\bibfield{author}{\bibinfo{person}{Prateep Bhattacharjee} {and}
  \bibinfo{person}{Sukhendu Das}.} \bibinfo{year}{2017}\natexlab{}.
\newblock \showarticletitle{Temporal Coherency based Criteria for Predicting
  Video Frames using Deep Multi-stage Generative Adversarial Networks}. In
  \bibinfo{booktitle}{{\em Advances in Neural Information Processing Systems}}.
  \bibinfo{pages}{4271--4280}.
\newblock


\bibitem[\protect\citeauthoryear{Bishop}{Bishop}{2006}]%
        {bishop2006book}
\bibfield{author}{\bibinfo{person}{Christopher~M. Bishop}.}
  \bibinfo{year}{2006}\natexlab{}.
\newblock \bibinfo{booktitle}{{\em Pattern Recognition and Machine Learning
  (Information Science and Statistics)}}.
\newblock \bibinfo{publisher}{Springer-Verlag New York, Inc.},
  \bibinfo{address}{Secaucus, NJ, USA}.
\newblock
\showISBNx{0387310738}


\bibitem[\protect\citeauthoryear{Chaitanya, Kaplanyan, Schied, Salvi, Lefohn,
  Nowrouzezahrai, and Aila}{Chaitanya et~al\mbox{.}}{2017}]%
        {chaitanya2017interactive}
\bibfield{author}{\bibinfo{person}{Chakravarty~Alla Chaitanya},
  \bibinfo{person}{Anton Kaplanyan}, \bibinfo{person}{Christoph Schied},
  \bibinfo{person}{Marco Salvi}, \bibinfo{person}{Aaron Lefohn},
  \bibinfo{person}{Derek Nowrouzezahrai}, {and} \bibinfo{person}{Timo Aila}.}
  \bibinfo{year}{2017}\natexlab{}.
\newblock \showarticletitle{Interactive reconstruction of Monte Carlo image
  sequences using a recurrent denoising autoencoder}.
\newblock \bibinfo{journal}{{\em ACM Transactions on Graphics (TOG)\/}}
  \bibinfo{volume}{36}, \bibinfo{number}{4} (\bibinfo{year}{2017}),
  \bibinfo{pages}{98}.
\newblock


\bibitem[\protect\citeauthoryear{Chen, Liao, Yuan, Yu, and Hua}{Chen
  et~al\mbox{.}}{2017}]%
        {Chen2017ICCV}
\bibfield{author}{\bibinfo{person}{Dongdong Chen}, \bibinfo{person}{Jing Liao},
  \bibinfo{person}{Lu Yuan}, \bibinfo{person}{Nenghai Yu}, {and}
  \bibinfo{person}{Gang Hua}.} \bibinfo{year}{2017}\natexlab{}.
\newblock \showarticletitle{Coherent Online Video Style Transfer}. In
  \bibinfo{booktitle}{{\em The IEEE International Conference on Computer Vision
  (ICCV)}}.
\newblock


\bibitem[\protect\citeauthoryear{Chu and Thuerey}{Chu and Thuerey}{2017}]%
        {chu2017cnnpatch}
\bibfield{author}{\bibinfo{person}{Mengyu Chu} {and} \bibinfo{person}{Nils
  Thuerey}.} \bibinfo{year}{2017}\natexlab{}.
\newblock \showarticletitle{Data-Driven Synthesis of Smoke Flows with
  {CNN}-based Feature Descriptors}.
\newblock \bibinfo{journal}{{\em {ACM} Trans. Graph.\/}}
  \bibinfo{volume}{36(4)}, \bibinfo{number}{69} (\bibinfo{year}{2017}).
\newblock


\bibitem[\protect\citeauthoryear{de~Bezenac, Pajot, and Gallinari}{de~Bezenac
  et~al\mbox{.}}{2017}]%
        {de2017deep}
\bibfield{author}{\bibinfo{person}{Emmanuel de Bezenac},
  \bibinfo{person}{Arthur Pajot}, {and} \bibinfo{person}{Patrick Gallinari}.}
  \bibinfo{year}{2017}\natexlab{}.
\newblock \showarticletitle{Deep Learning for Physical Processes: Incorporating
  Prior Scientific Knowledge}.
\newblock \bibinfo{journal}{{\em arXiv preprint arXiv:1711.07970\/}}
  (\bibinfo{year}{2017}).
\newblock


\bibitem[\protect\citeauthoryear{Dong, Loy, He, and Tang}{Dong
  et~al\mbox{.}}{2016}]%
        {dong2016image}
\bibfield{author}{\bibinfo{person}{Chao Dong}, \bibinfo{person}{Chen~Change
  Loy}, \bibinfo{person}{Kaiming He}, {and} \bibinfo{person}{Xiaoou Tang}.}
  \bibinfo{year}{2016}\natexlab{}.
\newblock \showarticletitle{Image super-resolution using deep convolutional
  networks}.
\newblock \bibinfo{journal}{{\em IEEE transactions on pattern analysis and
  machine intelligence\/}} \bibinfo{volume}{38}, \bibinfo{number}{2}
  (\bibinfo{year}{2016}), \bibinfo{pages}{295--307}.
\newblock


\bibitem[\protect\citeauthoryear{Dosovitskiy and Brox}{Dosovitskiy and
  Brox}{2016}]%
        {dosovitskiy2016generating}
\bibfield{author}{\bibinfo{person}{Alexey Dosovitskiy} {and}
  \bibinfo{person}{Thomas Brox}.} \bibinfo{year}{2016}\natexlab{}.
\newblock \showarticletitle{Generating images with perceptual similarity
  metrics based on deep networks}. In \bibinfo{booktitle}{{\em Advances in
  Neural Information Processing Systems}}. \bibinfo{pages}{658--666}.
\newblock


\bibitem[\protect\citeauthoryear{Dosovitskiy, Fischer, Springenberg,
  Riedmiller, and Brox}{Dosovitskiy et~al\mbox{.}}{2016}]%
        {dosovitskiy2016discriminative}
\bibfield{author}{\bibinfo{person}{Alexey Dosovitskiy},
  \bibinfo{person}{Philipp Fischer}, \bibinfo{person}{Jost~Tobias
  Springenberg}, \bibinfo{person}{Martin Riedmiller}, {and}
  \bibinfo{person}{Thomas Brox}.} \bibinfo{year}{2016}\natexlab{}.
\newblock \showarticletitle{Discriminative unsupervised feature learning with
  exemplar convolutional neural networks}.
\newblock \bibinfo{journal}{{\em {IEEE} Trans. Pattern Analysis and Mach.
  Int.\/}} \bibinfo{volume}{38}, \bibinfo{number}{9} (\bibinfo{year}{2016}),
  \bibinfo{pages}{1734--1747}.
\newblock


\bibitem[\protect\citeauthoryear{Farimani, Gomes, and Pande}{Farimani
  et~al\mbox{.}}{2017}]%
        {farimani2017}
\bibfield{author}{\bibinfo{person}{Amir~Barati Farimani},
  \bibinfo{person}{Joseph Gomes}, {and} \bibinfo{person}{Vijay~S Pande}.}
  \bibinfo{year}{2017}\natexlab{}.
\newblock \showarticletitle{Deep Learning the Physics of Transport Phenomena}.
\newblock \bibinfo{journal}{{\em arXiv:1709.02432\/}} (\bibinfo{year}{2017}).
\newblock


\bibitem[\protect\citeauthoryear{Flynn, Neulander, Philbin, and Snavely}{Flynn
  et~al\mbox{.}}{2016}]%
        {flynn2016deepstereo}
\bibfield{author}{\bibinfo{person}{John Flynn}, \bibinfo{person}{Ivan
  Neulander}, \bibinfo{person}{James Philbin}, {and} \bibinfo{person}{Noah
  Snavely}.} \bibinfo{year}{2016}\natexlab{}.
\newblock \showarticletitle{DeepStereo: Learning to predict new views from the
  world's imagery}. In \bibinfo{booktitle}{{\em Proceedings of the IEEE
  Conference on Computer Vision and Pattern Recognition}}.
  \bibinfo{pages}{5515--5524}.
\newblock


\bibitem[\protect\citeauthoryear{Girshick, Donahue, Darrell, and
  Malik}{Girshick et~al\mbox{.}}{2014}]%
        {girshick2014rich}
\bibfield{author}{\bibinfo{person}{Ross Girshick}, \bibinfo{person}{Jeff
  Donahue}, \bibinfo{person}{Trevor Darrell}, {and} \bibinfo{person}{Jitendra
  Malik}.} \bibinfo{year}{2014}\natexlab{}.
\newblock \showarticletitle{Rich feature hierarchies for accurate object
  detection and semantic segmentation}. In \bibinfo{booktitle}{{\em Proc. of
  {IEEE} Comp. Vision and Pattern Rec.}} IEEE, \bibinfo{pages}{580--587}.
\newblock


\bibitem[\protect\citeauthoryear{Goodfellow}{Goodfellow}{2016}]%
        {goodfellow2016nips}
\bibfield{author}{\bibinfo{person}{Ian Goodfellow}.}
  \bibinfo{year}{2016}\natexlab{}.
\newblock \showarticletitle{NIPS 2016 tutorial: Generative adversarial
  networks}.
\newblock \bibinfo{journal}{{\em arXiv preprint arXiv:1701.00160\/}}
  (\bibinfo{year}{2016}).
\newblock


\bibitem[\protect\citeauthoryear{Goodfellow, Bengio, and Courville}{Goodfellow
  et~al\mbox{.}}{2016}]%
        {Goodfellow2016}
\bibfield{author}{\bibinfo{person}{Ian Goodfellow}, \bibinfo{person}{Yoshua
  Bengio}, {and} \bibinfo{person}{Aaron Courville}.}
  \bibinfo{year}{2016}\natexlab{}.
\newblock \bibinfo{booktitle}{{\em Deep Learning}}.
\newblock \bibinfo{publisher}{MIT Press}.
\newblock


\bibitem[\protect\citeauthoryear{Goodfellow, Pouget-Abadie, Mirza, Xu,
  Warde-Farley, Ozair, Courville, and Bengio}{Goodfellow et~al\mbox{.}}{2014}]%
        {goodfellow2014generative}
\bibfield{author}{\bibinfo{person}{Ian~J Goodfellow}, \bibinfo{person}{Jean
  Pouget-Abadie}, \bibinfo{person}{Mehdi Mirza}, \bibinfo{person}{Bing Xu},
  \bibinfo{person}{David Warde-Farley}, \bibinfo{person}{Sherjil Ozair},
  \bibinfo{person}{Aaron Courville}, {and} \bibinfo{person}{Yoshua Bengio}.}
  \bibinfo{year}{2014}\natexlab{}.
\newblock \showarticletitle{Generative Adversarial Nets}.
\newblock \bibinfo{journal}{{\em stat\/}}  \bibinfo{volume}{1050}
  (\bibinfo{year}{2014}), \bibinfo{pages}{10}.
\newblock


\bibitem[\protect\citeauthoryear{Isola, Zhu, Zhou, and Efros}{Isola
  et~al\mbox{.}}{2017}]%
        {isola2016image}
\bibfield{author}{\bibinfo{person}{Phillip Isola}, \bibinfo{person}{Jun-Yan
  Zhu}, \bibinfo{person}{Tinghui Zhou}, {and} \bibinfo{person}{Alexei~A
  Efros}.} \bibinfo{year}{2017}\natexlab{}.
\newblock \showarticletitle{Image-to-image translation with conditional
  adversarial networks}.
\newblock \bibinfo{journal}{{\em Proc. of {IEEE} Comp. Vision and Pattern
  Rec.\/}} (\bibinfo{year}{2017}).
\newblock


\bibitem[\protect\citeauthoryear{Johnson, Alahi, and Fei-Fei}{Johnson
  et~al\mbox{.}}{2016}]%
        {johnson2016perceptual}
\bibfield{author}{\bibinfo{person}{Justin Johnson}, \bibinfo{person}{Alexandre
  Alahi}, {and} \bibinfo{person}{Li Fei-Fei}.} \bibinfo{year}{2016}\natexlab{}.
\newblock \showarticletitle{Perceptual losses for real-time style transfer and
  super-resolution}. In \bibinfo{booktitle}{{\em European Conference on
  Computer Vision}}. Springer, \bibinfo{pages}{694--711}.
\newblock


\bibitem[\protect\citeauthoryear{Kallweit, M{\"u}ller, McWilliams, Gross, and
  Nov{\'a}k}{Kallweit et~al\mbox{.}}{2017}]%
        {kallweit2017deep}
\bibfield{author}{\bibinfo{person}{Simon Kallweit}, \bibinfo{person}{Thomas
  M{\"u}ller}, \bibinfo{person}{Brian McWilliams}, \bibinfo{person}{Markus
  Gross}, {and} \bibinfo{person}{Jan Nov{\'a}k}.}
  \bibinfo{year}{2017}\natexlab{}.
\newblock \showarticletitle{Deep Scattering: Rendering Atmospheric Clouds with
  Radiance-Predicting Neural Networks}.
\newblock \bibinfo{journal}{{\em arXiv:1709.05418\/}} (\bibinfo{year}{2017}).
\newblock


\bibitem[\protect\citeauthoryear{Karras, Aila, Laine, and Lehtinen}{Karras
  et~al\mbox{.}}{2017}]%
        {karras2017growgan}
\bibfield{author}{\bibinfo{person}{Tero Karras}, \bibinfo{person}{Timo Aila},
  \bibinfo{person}{Samuli Laine}, {and} \bibinfo{person}{Jaakko Lehtinen}.}
  \bibinfo{year}{2017}\natexlab{}.
\newblock \showarticletitle{Progressive growing of gans for improved quality,
  stability, and variation}.
\newblock \bibinfo{journal}{{\em arXiv:1710.10196\/}} (\bibinfo{year}{2017}).
\newblock


\bibitem[\protect\citeauthoryear{Kavan, Gerszewski, Bargteil, and Sloan}{Kavan
  et~al\mbox{.}}{2011}]%
        {kavan2011physics}
\bibfield{author}{\bibinfo{person}{Ladislav Kavan}, \bibinfo{person}{Dan
  Gerszewski}, \bibinfo{person}{Adam~W Bargteil}, {and}
  \bibinfo{person}{Peter-Pike Sloan}.} \bibinfo{year}{2011}\natexlab{}.
\newblock \showarticletitle{Physics-inspired upsampling for cloth simulation in
  games}. In \bibinfo{booktitle}{{\em ACM Transactions on Graphics (TOG)}},
  Vol.~\bibinfo{volume}{30}. ACM, \bibinfo{pages}{93}.
\newblock


\bibitem[\protect\citeauthoryear{Kim, Liu, Llamas, and Rossignac}{Kim
  et~al\mbox{.}}{2005}]%
        {Kim05FlowFixer}
\bibfield{author}{\bibinfo{person}{Byungmoon Kim}, \bibinfo{person}{Yingjie
  Liu}, \bibinfo{person}{Ignacio Llamas}, {and} \bibinfo{person}{Jarek
  Rossignac}.} \bibinfo{year}{2005}\natexlab{}.
\newblock \showarticletitle{{FlowFixer: Using BFECC for Fluid Simulation}}. In
  \bibinfo{booktitle}{{\em Proceedings of the First Eurographics conference on
  Natural Phenomena}}. \bibinfo{pages}{51--56}.
\newblock


\bibitem[\protect\citeauthoryear{Kim, Kwon~Lee, and Mu~Lee}{Kim
  et~al\mbox{.}}{2016}]%
        {kim2016accurate}
\bibfield{author}{\bibinfo{person}{Jiwon Kim}, \bibinfo{person}{Jung Kwon~Lee},
  {and} \bibinfo{person}{Kyoung Mu~Lee}.} \bibinfo{year}{2016}\natexlab{}.
\newblock \showarticletitle{Accurate image super-resolution using very deep
  convolutional networks}. In \bibinfo{booktitle}{{\em Proceedings of the IEEE
  Conference on Computer Vision and Pattern Recognition}}.
  \bibinfo{pages}{1646--1654}.
\newblock


\bibitem[\protect\citeauthoryear{Kim, Thuerey, James, and Gross}{Kim
  et~al\mbox{.}}{2008}]%
        {Kim:2008:wlt}
\bibfield{author}{\bibinfo{person}{Theodore Kim}, \bibinfo{person}{Nils
  Thuerey}, \bibinfo{person}{Doug James}, {and} \bibinfo{person}{Markus
  Gross}.} \bibinfo{year}{2008}\natexlab{}.
\newblock \showarticletitle{{Wavelet Turbulence for Fluid Simulation}}.
\newblock \bibinfo{journal}{{\em {ACM} Trans. Graph.\/}}  \bibinfo{volume}{27
  (3)} (\bibinfo{year}{2008}), \bibinfo{pages}{50:1--6}.
\newblock


\bibitem[\protect\citeauthoryear{Krizhevsky, Sutskever, and Hinton}{Krizhevsky
  et~al\mbox{.}}{2012}]%
        {krizhevsky2012imagenet}
\bibfield{author}{\bibinfo{person}{Alex Krizhevsky}, \bibinfo{person}{Ilya
  Sutskever}, {and} \bibinfo{person}{Geoffrey~E Hinton}.}
  \bibinfo{year}{2012}\natexlab{}.
\newblock \showarticletitle{Imagenet classification with deep convolutional
  neural networks}. In \bibinfo{booktitle}{{\em Advances in Neural Information
  Processing Systems}}. NIPS, \bibinfo{pages}{1097--1105}.
\newblock


\bibitem[\protect\citeauthoryear{Ladicky, Jeong, Solenthaler, Pollefeys, and
  Gross}{Ladicky et~al\mbox{.}}{2015}]%
        {ladicky2015data}
\bibfield{author}{\bibinfo{person}{Lubor Ladicky}, \bibinfo{person}{SoHyeon
  Jeong}, \bibinfo{person}{Barbara Solenthaler}, \bibinfo{person}{Marc
  Pollefeys}, {and} \bibinfo{person}{Markus Gross}.}
  \bibinfo{year}{2015}\natexlab{}.
\newblock \showarticletitle{Data-driven fluid simulations using regression
  forests}.
\newblock \bibinfo{journal}{{\em {ACM} Trans. Graph.\/}} \bibinfo{volume}{34},
  \bibinfo{number}{6} (\bibinfo{year}{2015}), \bibinfo{pages}{199}.
\newblock


\bibitem[\protect\citeauthoryear{Ledig, Theis, Husz{\'a}r, Caballero,
  Cunningham, Acosta, Aitken, Tejani, Totz, Wang, et~al\mbox{.}}{Ledig
  et~al\mbox{.}}{2016}]%
        {ledig2016photo}
\bibfield{author}{\bibinfo{person}{Christian Ledig}, \bibinfo{person}{Lucas
  Theis}, \bibinfo{person}{Ferenc Husz{\'a}r}, \bibinfo{person}{Jose
  Caballero}, \bibinfo{person}{Andrew Cunningham}, \bibinfo{person}{Alejandro
  Acosta}, \bibinfo{person}{Andrew Aitken}, \bibinfo{person}{Alykhan Tejani},
  \bibinfo{person}{Johannes Totz}, \bibinfo{person}{Zehan Wang},
  {et~al\mbox{.}}} \bibinfo{year}{2016}\natexlab{}.
\newblock \showarticletitle{Photo-realistic single image super-resolution using
  a generative adversarial network}.
\newblock \bibinfo{journal}{{\em arXiv:1609.04802\/}} (\bibinfo{year}{2016}).
\newblock


\bibitem[\protect\citeauthoryear{Lim, Son, Kim, Nah, and Lee}{Lim
  et~al\mbox{.}}{2017}]%
        {lim2017enhanced}
\bibfield{author}{\bibinfo{person}{Bee Lim}, \bibinfo{person}{Sanghyun Son},
  \bibinfo{person}{Heewon Kim}, \bibinfo{person}{Seungjun Nah}, {and}
  \bibinfo{person}{Kyoung~Mu Lee}.} \bibinfo{year}{2017}\natexlab{}.
\newblock \showarticletitle{Enhanced deep residual networks for single image
  super-resolution}. In \bibinfo{booktitle}{{\em Proc. of {IEEE} Comp. Vision
  and Pattern Rec.}}, Vol.~\bibinfo{volume}{1}. \bibinfo{pages}{3}.
\newblock


\bibitem[\protect\citeauthoryear{Liu, Wang, Fan, Liu, Wang, Chang, and
  Huang}{Liu et~al\mbox{.}}{2017}]%
        {liu2017VideoSR}
\bibfield{author}{\bibinfo{person}{Ding Liu}, \bibinfo{person}{Zhaowen Wang},
  \bibinfo{person}{Yuchen Fan}, \bibinfo{person}{Xianming Liu},
  \bibinfo{person}{Zhangyang Wang}, \bibinfo{person}{Shiyu Chang}, {and}
  \bibinfo{person}{Thomas Huang}.} \bibinfo{year}{2017}\natexlab{}.
\newblock \showarticletitle{Robust Video Super-Resolution With Learned Temporal
  Dynamics}. In \bibinfo{booktitle}{{\em The IEEE International Conference on
  Computer Vision (ICCV)}}.
\newblock


\bibitem[\protect\citeauthoryear{Long, Lu, Ma, and Dong}{Long
  et~al\mbox{.}}{2017}]%
        {long2017pde}
\bibfield{author}{\bibinfo{person}{Zichao Long}, \bibinfo{person}{Yiping Lu},
  \bibinfo{person}{Xianzhong Ma}, {and} \bibinfo{person}{Bin Dong}.}
  \bibinfo{year}{2017}\natexlab{}.
\newblock \showarticletitle{PDE-Net: Learning PDEs from Data}.
\newblock \bibinfo{journal}{{\em arXiv:1710.09668\/}} (\bibinfo{year}{2017}).
\newblock


\bibitem[\protect\citeauthoryear{Luan, Paris, Shechtman, and Bala}{Luan
  et~al\mbox{.}}{2017}]%
        {luan2017deep}
\bibfield{author}{\bibinfo{person}{Fujun Luan}, \bibinfo{person}{Sylvain
  Paris}, \bibinfo{person}{Eli Shechtman}, {and} \bibinfo{person}{Kavita
  Bala}.} \bibinfo{year}{2017}\natexlab{}.
\newblock \showarticletitle{Deep Photo Style Transfer}.
\newblock \bibinfo{journal}{{\em arXiv preprint arXiv:1703.07511\/}}
  (\bibinfo{year}{2017}).
\newblock


\bibitem[\protect\citeauthoryear{Magnus, Henrik, Chris, and Stephen}{Magnus
  et~al\mbox{.}}{2011}]%
        {magnus2011capturing}
\bibfield{author}{\bibinfo{person}{W Magnus}, \bibinfo{person}{F Henrik},
  \bibinfo{person}{A Chris}, {and} \bibinfo{person}{M Stephen}.}
  \bibinfo{year}{2011}\natexlab{}.
\newblock \showarticletitle{Capturing Thin Features in Smoke Simulations}.
\newblock \bibinfo{journal}{{\em Siggraph Talk\/}} (\bibinfo{year}{2011}).
\newblock


\bibitem[\protect\citeauthoryear{Mathieu, Couprie, and LeCun}{Mathieu
  et~al\mbox{.}}{2015}]%
        {mathieu2015deep}
\bibfield{author}{\bibinfo{person}{Michael Mathieu}, \bibinfo{person}{Camille
  Couprie}, {and} \bibinfo{person}{Yann LeCun}.}
  \bibinfo{year}{2015}\natexlab{}.
\newblock \showarticletitle{Deep multi-scale video prediction beyond mean
  square error}.
\newblock \bibinfo{journal}{{\em arXiv preprint arXiv:1511.05440\/}}
  (\bibinfo{year}{2015}).
\newblock


\bibitem[\protect\citeauthoryear{McNamara, Treuille, Popovi\'{c}, and
  Stam}{McNamara et~al\mbox{.}}{2004}]%
        {McNamaraAdjointMethod}
\bibfield{author}{\bibinfo{person}{Antoine McNamara}, \bibinfo{person}{Adrien
  Treuille}, \bibinfo{person}{Zoran Popovi\'{c}}, {and} \bibinfo{person}{Jos
  Stam}.} \bibinfo{year}{2004}\natexlab{}.
\newblock \showarticletitle{{Fluid Control Using the Adjoint Method}}.
\newblock \bibinfo{journal}{{\em {ACM} Trans. Graph.\/}} \bibinfo{volume}{23},
  \bibinfo{number}{3} (\bibinfo{year}{2004}), \bibinfo{pages}{449--456}.
\newblock


\bibitem[\protect\citeauthoryear{Mirza and Osindero}{Mirza and
  Osindero}{2014}]%
        {mirza2014conditional}
\bibfield{author}{\bibinfo{person}{Mehdi Mirza} {and} \bibinfo{person}{Simon
  Osindero}.} \bibinfo{year}{2014}\natexlab{}.
\newblock \showarticletitle{Conditional generative adversarial nets}.
\newblock \bibinfo{journal}{{\em arXiv preprint arXiv:1411.1784\/}}
  (\bibinfo{year}{2014}).
\newblock


\bibitem[\protect\citeauthoryear{Mosser, Dubrule, and Blunt}{Mosser
  et~al\mbox{.}}{2017}]%
        {mosser2017porous}
\bibfield{author}{\bibinfo{person}{Lukas Mosser}, \bibinfo{person}{Olivier
  Dubrule}, {and} \bibinfo{person}{Martin~J Blunt}.}
  \bibinfo{year}{2017}\natexlab{}.
\newblock \showarticletitle{Reconstruction of three-dimensional porous media
  using generative adversarial neural networks}.
\newblock \bibinfo{journal}{{\em arXiv:1704.03225\/}} (\bibinfo{year}{2017}).
\newblock


\bibitem[\protect\citeauthoryear{Narain, Sewall, Carlson, and Lin}{Narain
  et~al\mbox{.}}{2008}]%
        {narain:2008:procTurb}
\bibfield{author}{\bibinfo{person}{Rahul Narain}, \bibinfo{person}{Jason
  Sewall}, \bibinfo{person}{Mark Carlson}, {and} \bibinfo{person}{Ming~C.
  Lin}.} \bibinfo{year}{2008}\natexlab{}.
\newblock \showarticletitle{{Fast Animation of Turbulence Using Energy
  Transport and Procedural Synthesis}}.
\newblock \bibinfo{journal}{{\em {ACM} Trans. Graph.\/}} \bibinfo{volume}{27},
  \bibinfo{number}{5} (\bibinfo{year}{2008}), \bibinfo{pages}{article 166}.
\newblock


\bibitem[\protect\citeauthoryear{Odena, Dumoulin, and Olah}{Odena
  et~al\mbox{.}}{2016}]%
        {odena2016deconv}
\bibfield{author}{\bibinfo{person}{Augustus Odena}, \bibinfo{person}{Vincent
  Dumoulin}, {and} \bibinfo{person}{Chris Olah}.}
  \bibinfo{year}{2016}\natexlab{}.
\newblock \showarticletitle{Deconvolution and Checkerboard Artifacts}.
\newblock \bibinfo{journal}{{\em Distill\/}} (\bibinfo{year}{2016}).
\newblock
\showDOI{%
\url{https://doi.org/10.23915/distill.00003}}


\bibitem[\protect\citeauthoryear{Pan, Huang, Tong, Zheng, and Bao}{Pan
  et~al\mbox{.}}{2013}]%
        {Pan:2013}
\bibfield{author}{\bibinfo{person}{Zherong Pan}, \bibinfo{person}{Jin Huang},
  \bibinfo{person}{Yiying Tong}, \bibinfo{person}{Changxi Zheng}, {and}
  \bibinfo{person}{Hujun Bao}.} \bibinfo{year}{2013}\natexlab{}.
\newblock \showarticletitle{Interactive Localized Liquid Motion Editing}.
\newblock \bibinfo{journal}{{\em {ACM} Trans. Graph.\/}} \bibinfo{volume}{32},
  \bibinfo{number}{6} (\bibinfo{date}{Nov.} \bibinfo{year}{2013}).
\newblock


\bibitem[\protect\citeauthoryear{Peng, Berseth, Yin, and Van De~Panne}{Peng
  et~al\mbox{.}}{2017}]%
        {peng2017deeploco}
\bibfield{author}{\bibinfo{person}{Xue~Bin Peng}, \bibinfo{person}{Glen
  Berseth}, \bibinfo{person}{KangKang Yin}, {and} \bibinfo{person}{Michiel Van
  De~Panne}.} \bibinfo{year}{2017}\natexlab{}.
\newblock \showarticletitle{{Deeploco: Dynamic locomotion skills using
  hierarchical deep reinforcement learning}}.
\newblock \bibinfo{journal}{{\em {ACM} Trans. Graph.\/}} \bibinfo{volume}{36},
  \bibinfo{number}{4} (\bibinfo{year}{2017}), \bibinfo{pages}{41}.
\newblock


\bibitem[\protect\citeauthoryear{Prantl, Bonev, and Thuerey}{Prantl
  et~al\mbox{.}}{2017}]%
        {rtliquids2017}
\bibfield{author}{\bibinfo{person}{Lukas Prantl}, \bibinfo{person}{Boris
  Bonev}, {and} \bibinfo{person}{Nils Thuerey}.}
  \bibinfo{year}{2017}\natexlab{}.
\newblock \showarticletitle{Pre-computed Liquid Spaces with Generative Neural
  Networks and Optical Flow}.
\newblock \bibinfo{journal}{{\em arXiv:1704.07854\/}} (\bibinfo{year}{2017}).
\newblock


\bibitem[\protect\citeauthoryear{Radford, Metz, and Chintala}{Radford
  et~al\mbox{.}}{2016}]%
        {RadfordMC15}
\bibfield{author}{\bibinfo{person}{Alec Radford}, \bibinfo{person}{Luke Metz},
  {and} \bibinfo{person}{Soumith Chintala}.} \bibinfo{year}{2016}\natexlab{}.
\newblock \showarticletitle{Unsupervised Representation Learning with Deep
  Convolutional Generative Adversarial Networks}.
\newblock \bibinfo{journal}{{\em Proc. ICLR\/}} (\bibinfo{year}{2016}).
\newblock


\bibitem[\protect\citeauthoryear{Rasmussen, Nguyen, Geiger, and
  Fedkiw}{Rasmussen et~al\mbox{.}}{2003}]%
        {rasmussen2003smoke}
\bibfield{author}{\bibinfo{person}{Nick Rasmussen}, \bibinfo{person}{Duc~Quang
  Nguyen}, \bibinfo{person}{Willi Geiger}, {and} \bibinfo{person}{Ronald
  Fedkiw}.} \bibinfo{year}{2003}\natexlab{}.
\newblock \showarticletitle{Smoke simulation for large scale phenomena}. In
  \bibinfo{booktitle}{{\em ACM Transactions on Graphics (TOG)}},
  Vol.~\bibinfo{volume}{22}. ACM, \bibinfo{pages}{703--707}.
\newblock


\bibitem[\protect\citeauthoryear{Ronneberger, Fischer, and Brox}{Ronneberger
  et~al\mbox{.}}{2015}]%
        {ronneberger2015u}
\bibfield{author}{\bibinfo{person}{Olaf Ronneberger}, \bibinfo{person}{Philipp
  Fischer}, {and} \bibinfo{person}{Thomas Brox}.}
  \bibinfo{year}{2015}\natexlab{}.
\newblock \showarticletitle{U-net: Convolutional networks for biomedical image
  segmentation}. In \bibinfo{booktitle}{{\em International Conference on
  Medical Image Computing and Computer-Assisted Intervention}}. Springer,
  \bibinfo{pages}{234--241}.
\newblock


\bibitem[\protect\citeauthoryear{Ruder, Dosovitskiy, and Brox}{Ruder
  et~al\mbox{.}}{2016}]%
        {Ruder2016Artistic}
\bibfield{author}{\bibinfo{person}{Manuel Ruder}, \bibinfo{person}{Alexey
  Dosovitskiy}, {and} \bibinfo{person}{Thomas Brox}.}
  \bibinfo{year}{2016}\natexlab{}.
\newblock \showarticletitle{Artistic Style Transfer for Videos}. In
  \bibinfo{booktitle}{{\em Pattern Recognition - 38th German Conference, {GCPR}
  2016, Hannover, Germany, September 12-15, 2016, Proceedings}}.
  \bibinfo{pages}{26--36}.
\newblock
\showDOI{%
\url{https://doi.org/10.1007/978-3-319-45886-1_3}}


\bibitem[\protect\citeauthoryear{Saito, Matsumoto, and Saito}{Saito
  et~al\mbox{.}}{2017}]%
        {saito2017temporal}
\bibfield{author}{\bibinfo{person}{Masaki Saito}, \bibinfo{person}{Eiichi
  Matsumoto}, {and} \bibinfo{person}{Shunta Saito}.}
  \bibinfo{year}{2017}\natexlab{}.
\newblock \showarticletitle{Temporal generative adversarial nets with singular
  value clipping}. In \bibinfo{booktitle}{{\em IEEE International Conference on
  Computer Vision (ICCV)}}. \bibinfo{pages}{2830--2839}.
\newblock


\bibitem[\protect\citeauthoryear{Salimans, Goodfellow, Zaremba, Cheung,
  Radford, and Chen}{Salimans et~al\mbox{.}}{2016}]%
        {salimans2016improved}
\bibfield{author}{\bibinfo{person}{Tim Salimans}, \bibinfo{person}{Ian
  Goodfellow}, \bibinfo{person}{Wojciech Zaremba}, \bibinfo{person}{Vicki
  Cheung}, \bibinfo{person}{Alec Radford}, {and} \bibinfo{person}{Xi Chen}.}
  \bibinfo{year}{2016}\natexlab{}.
\newblock \showarticletitle{Improved techniques for training gans}. In
  \bibinfo{booktitle}{{\em Advances in Neural Information Processing Systems}}.
  \bibinfo{pages}{2234--2242}.
\newblock


\bibitem[\protect\citeauthoryear{Schechter and Bridson}{Schechter and
  Bridson}{2008}]%
        {schechter2008evolving}
\bibfield{author}{\bibinfo{person}{Hagit Schechter} {and}
  \bibinfo{person}{Robert Bridson}.} \bibinfo{year}{2008}\natexlab{}.
\newblock \showarticletitle{Evolving sub-grid turbulence for smoke animation}.
  In \bibinfo{booktitle}{{\em Proceedings of the 2008 ACM SIGGRAPH/Eurographics
  symposium on Computer animation}}. Eurographics Association,
  \bibinfo{pages}{1--7}.
\newblock


\bibitem[\protect\citeauthoryear{Selle, Fedkiw, Kim, Liu, and Rossignac}{Selle
  et~al\mbox{.}}{2008}]%
        {Selle:2008:USM}
\bibfield{author}{\bibinfo{person}{Andrew Selle}, \bibinfo{person}{Ronald
  Fedkiw}, \bibinfo{person}{Byungmoon Kim}, \bibinfo{person}{Yingjie Liu},
  {and} \bibinfo{person}{Jarek Rossignac}.} \bibinfo{year}{2008}\natexlab{}.
\newblock \showarticletitle{{An Unconditionally Stable MacCormack Method}}.
\newblock \bibinfo{journal}{{\em J. Sci. Comput.\/}} \bibinfo{volume}{35},
  \bibinfo{number}{2-3} (\bibinfo{date}{June} \bibinfo{year}{2008}),
  \bibinfo{pages}{350--371}.
\newblock


\bibitem[\protect\citeauthoryear{Simonyan and Zisserman}{Simonyan and
  Zisserman}{2014}]%
        {simonyan2014}
\bibfield{author}{\bibinfo{person}{Karen Simonyan} {and}
  \bibinfo{person}{Andrew Zisserman}.} \bibinfo{year}{2014}\natexlab{}.
\newblock \showarticletitle{Very deep convolutional networks for large-scale
  image recognition}.
\newblock \bibinfo{journal}{{\em arXiv:1409.1556\/}} (\bibinfo{year}{2014}).
\newblock


\bibitem[\protect\citeauthoryear{Stam}{Stam}{1999}]%
        {Stam1999}
\bibfield{author}{\bibinfo{person}{Jos Stam}.} \bibinfo{year}{1999}\natexlab{}.
\newblock \showarticletitle{{Stable Fluids}}. In \bibinfo{booktitle}{{\em Proc.
  ACM SIGGRAPH}}. ACM, \bibinfo{pages}{121--128}.
\newblock


\bibitem[\protect\citeauthoryear{Tompson, Schlachter, Sprechmann, and
  Perlin}{Tompson et~al\mbox{.}}{2016}]%
        {tompson2016accelerating}
\bibfield{author}{\bibinfo{person}{Jonathan Tompson},
  \bibinfo{person}{Kristofer Schlachter}, \bibinfo{person}{Pablo Sprechmann},
  {and} \bibinfo{person}{Ken Perlin}.} \bibinfo{year}{2016}\natexlab{}.
\newblock \showarticletitle{Accelerating Eulerian Fluid Simulation With
  Convolutional Networks}.
\newblock \bibinfo{journal}{{\em arXiv: 1607.03597\/}} (\bibinfo{year}{2016}).
\newblock


\bibitem[\protect\citeauthoryear{Um, Hu, and Thuerey}{Um et~al\mbox{.}}{2017}]%
        {um2017mlflip}
\bibfield{author}{\bibinfo{person}{Kiwon Um}, \bibinfo{person}{Xiangyu Hu},
  {and} \bibinfo{person}{Nils Thuerey}.} \bibinfo{year}{2017}\natexlab{}.
\newblock \showarticletitle{Splash Modeling with Neural Networks}.
\newblock \bibinfo{journal}{{\em arXiv:1704.04456\/}} (\bibinfo{year}{2017}).
\newblock


\bibitem[\protect\citeauthoryear{Yu, Zhang, Wang, and Yu}{Yu
  et~al\mbox{.}}{2017}]%
        {yu2017seqgan}
\bibfield{author}{\bibinfo{person}{Lantao Yu}, \bibinfo{person}{Weinan Zhang},
  \bibinfo{person}{Jun Wang}, {and} \bibinfo{person}{Yong Yu}.}
  \bibinfo{year}{2017}\natexlab{}.
\newblock \showarticletitle{SeqGAN: Sequence Generative Adversarial Nets with
  Policy Gradient.}. In \bibinfo{booktitle}{{\em AAAI}}.
  \bibinfo{pages}{2852--2858}.
\newblock


\bibitem[\protect\citeauthoryear{Zhao, Gallo, Frosio, and Kautz}{Zhao
  et~al\mbox{.}}{2015}]%
        {zhao2015loss}
\bibfield{author}{\bibinfo{person}{Hang Zhao}, \bibinfo{person}{Orazio Gallo},
  \bibinfo{person}{Iuri Frosio}, {and} \bibinfo{person}{Jan Kautz}.}
  \bibinfo{year}{2015}\natexlab{}.
\newblock \showarticletitle{Loss Functions for Neural Networks for Image
  Processing}.
\newblock \bibinfo{journal}{{\em arXiv preprint arXiv:1511.08861\/}}
  (\bibinfo{year}{2015}).
\newblock


\bibitem[\protect\citeauthoryear{Zhu, Park, Isola, and Efros}{Zhu
  et~al\mbox{.}}{2017}]%
        {zhu2017cycle}
\bibfield{author}{\bibinfo{person}{Jun-Yan Zhu}, \bibinfo{person}{Taesung
  Park}, \bibinfo{person}{Phillip Isola}, {and} \bibinfo{person}{Alexei~A
  Efros}.} \bibinfo{year}{2017}\natexlab{}.
\newblock \showarticletitle{Unpaired image-to-image translation using
  cycle-consistent adversarial networks}.
\newblock \bibinfo{journal}{{\em arXiv:1703.10593\/}} (\bibinfo{year}{2017}).
\newblock


\end{thebibliography}

\appendix
\footnotesize
\section{Details of NN Architectures}
\label{app:nnarch}

To clearly specify our networks, we use the following notation. Let $in$(resolution, channels), $out$(resolution, output) present input and output information; $NI$(output-resolution) represent nearest-neighbor interpolation; $C$(output-resolution, filter size, output-channels) denote a convolutional layer.
Our resolutions and filter sizes are the same for every spacial dimension for both 2D and 3D.
Resolutions of feature maps are reduced when strides $>$1. 
We use $RB$ to represent our residual blocks, and use $C_S$
for adding residuals in a $RB$. 
E.g., $RB_3: [C_A, \text{ReLU}, C_B]+[C_S], \text{ReLU}$ means $[(input \rightarrow C_A \rightarrow	\text{ReLU} \rightarrow C_B)+(input \rightarrow C_S)] \rightarrow	\text{ReLU}$, where + denotes element-wise addition. $BN$ denotes batch normalization, which is not used in the last layer of $G$, the first layer of $D_t$ and the first layer of $D_s$ \cite{RadfordMC15}. In addition, | denotes concatenation of layer outputs along the channel dimension.\\
\\
Architectures of $G$, $D_s$ and $D_t$:\\
\resizebox{\hsize}{!}{\renewcommand{\arraystretch}{1.1}
	\begin{tabular}{|l|}
	\hline $G$:\\ \hline
	$in(16,4)$                                          \\ \hline
	$NI(64,4)$                                            \\ \hline
	$RB_0: [C_A(64,5,8), BN, \text{ReLU}, C_B(64,5,32), BN]+[C_S(64,1,32), BN], \text{ReLU}$     \\ \hline
	$RB_1: [C_A(64,5,128), BN, \text{ReLU}, C_B(64,5,128), BN]+[C_S(64,1,128), BN], \text{ReLU}$ \\ \hline
	$RB_2: [C_A(64,5,32), BN, \text{ReLU}, C_B(64,5,8), BN]+[C_S(64,1,8), BN], \text{ReLU}$      \\ \hline
	$RB_3: [C_A(64,5,2), \text{ReLU}, C_B(64,5,1)]+[C_S(64,1,1)], \text{ReLU}$       \\ \hline
	$out(64,1)$       \\ \hline
	\end{tabular}
}\\
\\
\resizebox{\hsize}{!}{\renewcommand{\arraystretch}{1.15}
	\begin{tabular}{|l|l|}
		\hline
		$D_s$: & $D_t$:\\ \hline
		$in_x(16,1)$        , the conditional density &	
		\multirow{ 2}{*}{$in_y(64,3)$, the 3 high-res frames to classify}\\
		$NI(64,1) | in_y(64,1)$           , the high-res input to classify  & \\ \hline
		$C(32, 4, 32), \text{leaky ReLU}$  & $C(32, 4, 32), \text{leaky ReLU}$   \\ \hline
		$C(16, 4, 64), BN, \text{leaky ReLU}$ & $C(16, 4, 64), BN, \text{leaky ReLU}$\\ \hline
		$C(8, 4, 128), BN, \text{leaky ReLU}$  &  $C(8, 4, 128), BN, \text{leaky ReLU}$ \\ \hline
		$C(8, 4, 256), BN, \text{leaky ReLU}$   &  $C(8, 4, 256), BN, \text{leaky ReLU}$   \\ \hline
		Fully connected, $\sigma$ activation   & Fully connected, $\sigma$ activation  \\ \hline
		$out(1, 1)$   &  $out(1, 1)$   \\ \hline
	\end{tabular}
}
\section{Parameters \& Data Statistics}
\label{app:data}

Below we summarize all parameters for training runs and data generation. Note
that the model size includes compression, and we train the individual networks
multiple times per iteration, as indicated below.\\
\\
Details of generated results:\\
	\resizebox{\hsize}{!}{
	\begin{tabular}{|l|l|l|l|l|}
		\hline
		test       & input size & \begin{tabular}[c]{@{}l@{}}tile ($34^{3}$) \\number\end{tabular} & output size & time            \\ \hline
		\myreffig{fig:styleLLCompare} a)         & $128^{2}$            & -                                 & $512^{2}$                 & 0.064s/frame     \\ \hline
		-          & $34^{3}$            & 1                                 & $136^{3}$                 & 2.2s/frame     \\ \hline
		\myreffig{fig:plume3d} b)     & $64^{3}$            & 8                                 & $256^{3}$                 & 17.9s/frame    \\ \hline
		\myreffig{fig:green} & $150\times{100}\times{100}$         & 96                                & $600\times{400}\times{400}$                 & 234.48s/frame  \\ \hline
		\myreffig{fig:house} & $256\times{180}\times{180}$     & 441                               & $1024\times{720}\times{720}$                & 1046.07s/frame~ \\ \hline
	\end{tabular}
	}
\\
\\
\\
Details of training runs for different models are listed in the following table. 
Our standard models that are used unless otherwise indicated are marked with a $^{(*)}$:
\vspace{2pt}
\\
\resizebox{\hsize}{!}{
	\begin{tabular}{|c|c|c|c|c|c|}
		\hline
		Train. iters & \begin{tabular}[c]{@{}l@{}}data: no. of sims,\\  total frames\end{tabular} &
		\begin{tabular}[c]{@{}l@{}}training and \\testing frames\end{tabular} &
		\begin{tabular}[c]{@{}l@{}}low-\\res\end{tabular}, \begin{tabular}[c]{@{}l@{}}high-\\res\end{tabular}, \begin{tabular}[c]{@{}l@{}}input\\tiles\end{tabular} &
		$\lambda_{L_1}$ &
		$\lambda_{f}^{1,...,4}$ 
		\\\hline
		2D, \myreffig{fig:styleLLCompare} a)$^{(*)}$ &  20, 4000  & 160, 40 &
		\multirow{3}{*}{$64^2,256^2,16^2$} &
		\multirow{3}{*}{5} & $-10^{-5} $ for all \Tstrut
		\\ \cline{1-3} \cline{6-6}
		2D, \myreffig{fig:styleLLCompare} b) &  20, 4000  & 160, 40 &
		&  & \begin{tabular}[c]{@{}l@{}} $10^{-4}/3,-10^{-4}/3,$\\$-10^{-4}/3,10^{-4}/3$ \end{tabular}\Tstruta
		\\ \cline{1-3} \cline{6-6}
		2D, \myreffig{fig:waveletTurbulence} b) &  20, 2400  & 320, 80 &
		&  & $10^{-5}$ for all \Tstrut
		\\ \hline
		3D, \myreffig{fig:plume3d} b)$^{(*)}$ &
		20, 2400  & 96, 24 & 
		\multirow{2}{*}{$64^3,256^3,16^3$} &
		\multirow{2}{*}{5} & $-10^{-6}/3 $ for all \Tstrut
		\\ \cline{1-3} \cline{6-6} 
		3D, \myreffig{fig:3ddifferentstyle} b) &
		20, 2400  &  96, 24 &
		&  & \begin{tabular}[c]{@{}l@{}} $10^{-6}/3,-10^{-6}/3,$\\$-10^{-6}/3,-10^{-6}/3$ \end{tabular}\Tstruta
		\\ \hline 
		\hline
		Table Cont. & 
		\begin{tabular}[c]{@{}l@{}}Trainings\\ per step\end{tabular} &
		\begin{tabular}[c]{@{}l@{}}Training \\ steps\end{tabular},
		\begin{tabular}[c]{@{}l@{}}Batch\\ size\end{tabular}&
		\begin{tabular}[c]{@{}l@{}}Model \\ weights\end{tabular} &
		\begin{tabular}[c]{@{}l@{}}Model\\ size (Mb)\end{tabular} &
		\begin{tabular}[c]{@{}l@{}}Training \\ time (min)\end{tabular}\\\hline
		2D, \myreffig{fig:styleLLCompare} a)$^{(*)}$ &  
		\multirow{3}{*}{\begin{tabular}[c]{@{}l@{}}2 for $D_s$, \\ 2 for $D_t$,\\ 2 for $G$\end{tabular}} &
		\multirow{3}{*}{40k, 16} &
		\multirow{3}{*}{\begin{tabular}[c]{@{}l@{}} $G$, 634214\\$D_s$, 706017\\ $D_t$, 706529\end{tabular}} &
		36.88 & 798.65 \\ \cline{1-1} \cline{5-6} 
		2D, \myreffig{fig:styleLLCompare} b) & &
		&  &  
		42.73 & 905.72 \\ \cline{1-1} \cline{5-6} 
		2D, \myreffig{fig:waveletTurbulence} b) & &
		&  & 
		43.45 & 877.59 \\ \hline
		\begin{tabular}[c]{@{}l@{}}3D, \myreffig{fig:plume3d} b)$^{(*)}$\end{tabular} & 
		\multirow{2}{*}{\begin{tabular}[c]{@{}l@{}}16 for $D_s$, \\ 16 for $D_t$,\\ 16 for $G$\end{tabular}} &
		\multirow{2}{*}{7k, 1} & 
		\multirow{2}{*}{\begin{tabular}[c]{@{}l@{}}$G$: 3148014\\ $D_s$: 2888161\\ $D_t$: 2890209\end{tabular}}
		& 134.93 & 
		\begin{tabular}[c]{@{}l@{}}12636.22\\ 2 GPUs\end{tabular} 
		\\ \cline{1-1}  \cline{5-6} 
		\begin{tabular}[c]{@{}l@{}}3D, \myreffig{fig:3ddifferentstyle} b)\end{tabular} &
		&  &  
		& 140.79 &
		\begin{tabular}[c]{@{}l@{}}18231.97\\ 2 GPUs\end{tabular} \\ \hline 
	\end{tabular}
}\\

\end{document}